%% file: pact_main.tex
\documentclass[sigconf,nonacm]{acmart}

\input{packages}
\input{commands}
\AtBeginDocument{%
  \providecommand\BibTeX{{%
    \normalfont B\kern-0.5em{\scshape i\kern-0.25em b}\kern-0.8em\TeX}}}

\acmSubmissionID{520}

\begin{document}

\acmYear{2022}\copyrightyear{2022}
\setcopyright{acmcopyright}
\acmConference[PACT '22]{International Conference on Parallel Architectures and Compilation Techniques}{October 10--12, 2022}{Chicago, IL, USA}
\acmBooktitle{International Conference on Parallel Architectures and Compilation Techniques (PACT '22), October 10--12, 2022, Chicago, IL, USA}
\acmPrice{15.00}
\acmDOI{10.1145/3559009.3569693}
\acmISBN{978-1-4503-9868-8/22/10}

%%
%% Cover letter for PACT Revision
% \input{index}

\title[Slice-and-Forge]{Slice-and-Forge: Making Better Use of Caches for \\ Graph Convolutional Network Accelerators\vspace{5mm}} 

\settopmatter{authorsperrow=4}
\author{Mingi Yoo}
\authornotemark[1]
\orcid{0000-0003-0215-5092}
%\authornote{Equal contribution}
\affiliation{%
\institution{Yonsei University}%
\city{Seoul}\country{South Korea}}
\email{skys7297@yonsei.ac.kr}
\author{Jaeyong Song}
\authornotemark[1]
\orcid{0000-0001-9976-7487}
\affiliation{%
\institution{Yonsei University}%
\city{Seoul}\country{South Korea}}
\email{jaeyong.song@yonsei.ac.kr}
\author{Hyeyoon Lee}
\orcid{0000-0003-3130-7921}
\affiliation{%
\institution{Yonsei University}%
\city{Seoul}\country{South Korea}}
\email{hylee817@yonsei.ac.kr}
\author{Jounghoo Lee}
\orcid{0000-0002-0463-7717}
\affiliation{%
\institution{Yonsei University}%
\city{Seoul}\country{South Korea}}
\email{jounghoolee@yonsei.ac.kr}
\author{Namhyung Kim}
\orcid{0000-0002-2030-6010}
\affiliation{%
\institution{Samsung Electronics}%
\city{Hwaseong}\country{South Korea}}
\email{namhyungk11@gmail.com}
\author{Youngsok Kim}
\orcid{0000-0002-1015-9969}
\affiliation{%
\institution{Yonsei University}%
\city{Seoul}\country{South Korea}}
\email{youngsok@yonsei.ac.kr}
\author{Jinho Lee}
\orcid{0000-0003-4010-6611}
%\authornote{Corresponding author}
\authornotemark[2]
\affiliation{%
\institution{Seoul National University}%
\city{Seoul}\country{South Korea}}
\email{leejinho@snu.ac.kr}

\thanks{\\ \\ This is an author preprint version of a paper which will appear in the proceedings of PACT'22.\\
$^*$ Equal contribution \\
$^{\dagger}$ Corresponding author.\\ \\ \\ }


\begin{abstract}
\vspace{3mm}
Graph convolutional networks (GCNs) are becoming increasingly popular as they can process a wide variety of data formats that prior deep neural networks cannot easily support. 
One key challenge in designing hardware accelerators for GCNs is the vast size and randomness in their data access patterns which greatly reduces the effectiveness of the limited on-chip cache.
Aimed at improving the effectiveness of the cache by mitigating the irregular data accesses, prior studies often employ the vertex tiling techniques used in traditional graph processing applications.
While being effective at enhancing the cache efficiency, those approaches are often sensitive to the tiling configurations where the optimal setting heavily depends on target input datasets.
Furthermore, the existing solutions require manual tuning through trial-and-error or rely on sub-optimal analytical models.

In this paper, we propose \emph{Slice-and-Forge (\accname)}, an efficient hardware accelerator for GCNs which greatly improves the effectiveness of the limited on-chip cache.
\accname chooses a tiling strategy named \fes that splits the features into vertical slices and processes them in the outermost loop of the execution. 
This particular choice results in a repetition of the identical computational patterns over irregular graph data over multiple rounds.
Taking advantage of such repetitions, \accname dynamically tunes its tile size.
Our experimental results reveal that \accname can achieve 1.73$\times$ higher performance in geomean compared to prior work on multi-engine settings, and 1.46$\times$ higher performance in geomean on small scale settings, without the need for off-line analyses.
\end{abstract}

% %%
% %% The code below is generated by the tool at http://dl.acm.org/ccs.cfm.
% %% Please copy and paste the code instead of the example below.
% %%
\begin{CCSXML}
<ccs2012>
   <concept>
       <concept_id>10010520.10010521.10010542.10010294</concept_id>
       <concept_desc>Computer systems organization~Neural networks</concept_desc>
       <concept_significance>500</concept_significance>
       </concept>
   <concept>
       <concept_id>10010520.10010521.10010528.10010534</concept_id>
       <concept_desc>Computer systems organization~Single instruction, multiple data</concept_desc>
       <concept_significance>500</concept_significance>
       </concept>
 </ccs2012>
\end{CCSXML}

\ccsdesc[500]{Computer systems organization~Neural networks}
\ccsdesc[500]{Computer systems organization~Single instruction, multiple data}

\keywords{\\ Graph Convolutional Networks, Accelerators, Caches}

\maketitle

% \newpage
%\vspace{3mm}
\section{Introduction}

\vspace{3mm}
%\JL{more about CNNS?}
% \JL{change access back to accesses}
% \JL{that -> which, even though -> although} 
% \JL{accesses countable?}
% After the ground-breaking success of convolutional neural networks in the image classification field~\cite{alexnet, resnet}, many types of deep neural networks (DNNs) have been introduced.
% Commonly used networks include recurrent neural networks~\cite{rnn}, generative adversarial networks~\cite{gan, stylegan}, and transformers~\cite{transformer}.

Graph convolutional networks (GCNs) are considered the next-generation deep neural networks (DNNs) as they can be applied to a variety of data which exhibit irregular patterns.
Despite their wide applicability spanning from traditional graph problems~\cite{edgepred,molecule} to modern DNN tasks~\cite{graphnlp}, conventional DNN accelerators cannot efficiently accelerate GCNs.
Different from traditional DNNs, % which comprise dense matrix operations, 
graphs exhibit highly sparse structures and demand fundamentally different acceleration schemes.
Although some hardware accelerators which exploit the sparsity in DNNs~\cite{Parashar2017SCNN, Judd2017Cnvlutin2, Zhang2016CambriconX} may cope with the highly sparse structures of graphs, such accelerators are still not well-suited for GCNs as the portion of non-zero values in graphs ($<$0.01\%) is significantly lower than sparse DNNs (10\%--90\%).

Aimed at the efficient acceleration of GCNs, prior studies propose new hardware accelerators which exploit various unique characteristics of GCNs. % when compared to traditional DNNs.
In particular, as GCNs incur large working set sizes and highly irregular memory access patterns, the hardware accelerators should efficiently utilize their limited on-chip caches. % to achieve high performance.
%For example, HyGCN\mbox{\cite{hygcn}} exploits the hybrid nature of GCN workloads by integrating SIMD units with systolic arrays.
%In addition, AWB-GCN\mbox{\cite{awb}} identifies a significant load imbalance as the bottleneck for GCN execution and proposes multi level load balancing techniques.
%Further, EnGN\mbox{\cite{engn}} and GCNAX\mbox{\cite{gcnax}} optimize dataflows for efficient data loading and computation.
Viewing GCNs as cascaded sparse-dense matrix multiplications, one popular technique is loop tiling and reordering, as demonstrated in~\cite{gcnax, engn, awb}. %\emph{vertex tiling}.
Loop tiling partitions a given GCN into multiple tiles, as performed in sparse matrix-vector multiplication (SpMV) approaches~\cite{gridgraph, spmm3}.
By confining the working set to a specific region, the loop tiling result in better cache efficiency at the expense of increased repetition counts.

% \begin{figure*}
%  \includegraphics[width=0.7\textwidth]{figs/intro_nolac.pdf}
%  \caption{Conceptual diagram of the \saf scheme.}
%  \label{fig:intro}
% \end{figure*}

However, the difficulty of loop tiling lies in the fact that configuring the tile size is non-trivial.
Especially for GCNs which exhibit abundant random accesses, 
the optimal setting for tiling varies per dataset and cannot be decided statically.
Thus, the current existing solution to obtain the optimal tiling configuration is manual tuning through trial-and-error.
While there exists an approach to restrictively apply analytical search for tiling configuration~\cite{gcnax}, it ignores the fact that the graph is sparse and often falls into an inferior solution.
% Do I need an example here?

\begin{figure}
 \includegraphics[width=\columnwidth]{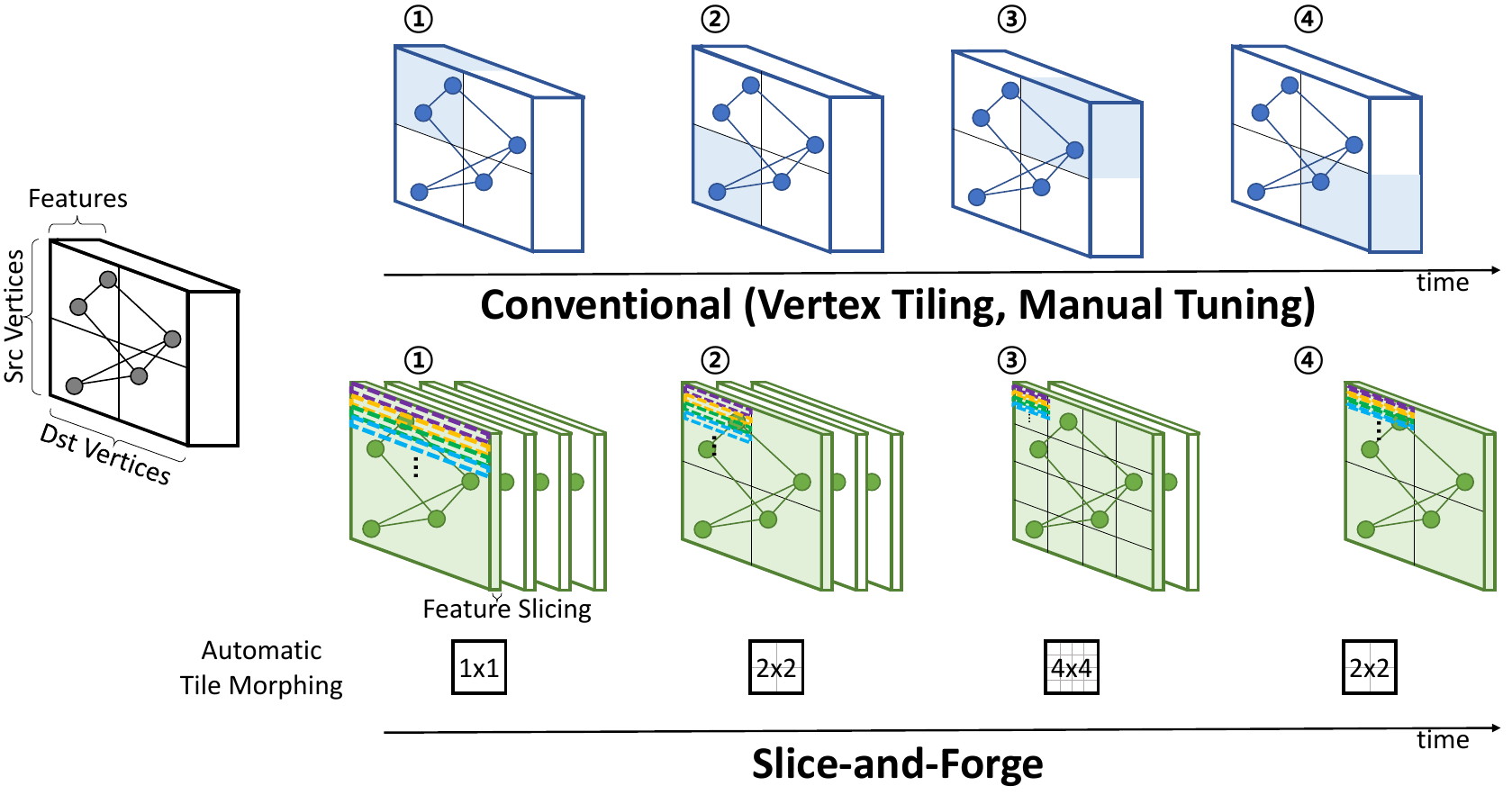}
 \caption{Conceptual diagram of the \saf scheme.}
 %\vspace{0mm}
 \label{fig:intro}
\end{figure}

In this work, we propose \emph{\SaF (\accname)}, a technique for enabling on-line tuning for the GCN configuration. % by exploiting repetitive access patterns.
The proposed scheme is conceptually illustrated in \fig{fig:intro}.
\accname sits on top of \emph{\fes}, which splits the feature matrix into multiple vertical strips and puts its loop at the outermost level.
This creates a repeated access pattern over the random graphs and allows for gradually tuning the configurations such that it can settle at a near-optimal configuration within a few rounds.

Exploiting the repetitive patterns, we propose a method for efficiently tuning the vertex tiling configurations with \emph{\atm}.
In vertex tiling, the adjacency matrix is partitioned into multiple tiles where the performance is sensitive to tile size.
With \atm, we search for the optimal tile size using a method similar to a convex optimization.
% Second is \emph{\lac} that determines how to distribute the rows in a tile among multiple execution engines such that their assembled access pattern exhibit better hit ratio. 
% The tuning is also performed in a similar manner in conjunction with \atm. 

Although the \fes itself is a kind of loop tiling technique that has been long studied in the context of SpMM~\cite{spmm1,spmm2} or GCNs~\cite{gcnax}, exploiting the repetitive pattern from \fes for dynamic configuration has not been studied yet, to the best of our knowledge.
In addition to enabling the repetitive dynamic configuration, we also show that \fes itself is often a favorable choice for loop tiling.
%, such that this choice itself often results in much higher speedups.
%somewhere: tuning sensitivity?

Our contributions can be summarized as follows:
\begin{itemize}[noitemsep,nolistsep,leftmargin=*]
    %\item We propose \SaF (\accname), a hardware accelerator that automatically tunes the tiling configuration over repetitive execution patterns.
    \item We identify that \fes creates repetitive patterns over a random graph data and propose \SaF (\accname) which exploits the pattern for efficient dynamic tuning.
    %\item Taking advantage of the fact that \fes incurs repeating computational patterns, 
    %    we propose performing locality optimization during repetitions.
    \item We suggest \atm that dynamically tunes the number of vertex tiles at runtime. % and \lac
    \item \rev{We provide a strategy for extending \accname to multi-chip module scaling.}
    \item We provide a thorough evaluation of \accname which composes the above techniques, and we discuss the benefits.
\end{itemize}

\section{Background}
\subsection{Graph Convolutional Networks}
%\JL{have to be careful. I copied this whole section to SGCN}
\label{sec:gcnbackground}
%\JL{have to be careful. I copied this whole section to SGCN}
\label{sec:gcnbackground}

\begin{figure}
\centering
 \includegraphics[width=\columnwidth]{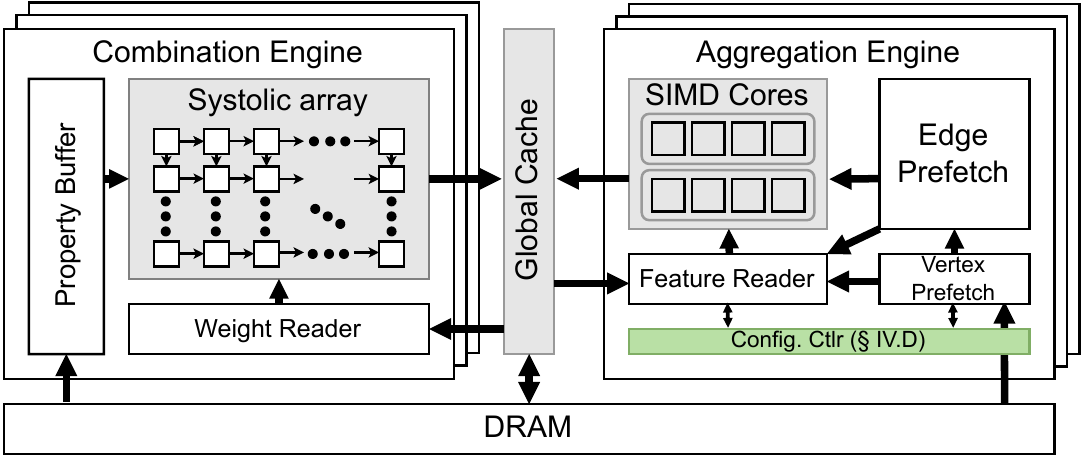}
 % \vspace{-3mm}
 \caption{Baseline GCN accelerator architecture.}
% \vspace{-3mm}
 \label{fig:accelerator}
\end{figure}

With numerous variations~\cite{gcn, graphsage, ginconv}, inference of a GCN layer $l$ can be formulated as follows:

\begin{equation}
\label{eq:gcn}
    X^{(l+1)} = \sigma(\tilde{A}\cdot X^{(l)} \cdot W^{(l)}).
\end{equation}
In the above equation, $\tilde{A} \in \mathbb{R}^{|V|\times|V|}$ is a normalized Laplacian\footnote{Normalized Laplacian refers to a form of adjacency matrix defined as $\tilde{A}=I-D^{-1/2}AD^{-1/2}$ where $D$ is a degree matrix.}. 
$X^{(l)} \in \mathbb{R}^{|V|\times |F^{(l)}|}$ is a feature matrix for the $l$-th layer where each row represents the features associated with a vertex.
$W^{(l)} \in \mathbb{R}^{|F^{(l)}| \times |F^{(l+1)}|}$ is a trainable weight matrix for layer $l$ and is the trainable part of GCNs.
$\sigma(\cdot)$ is an activation function where the rectified linear unit (ReLU) is commonly used. % for $\sigma(\cdot)$.
Typically, three to five layers comprise a GCN~\cite{deepgcn, deeper} as having too many layers is known to be difficult to train.

\cref{eq:gcn} has two matrix multiplications in it.
The first, matrix multiplication between \A and $X^{(l)}$ or \A and $X^{(l)} \cdot W^{(l)}$, is called the \emph{aggregation} phase.
Because each row of \A represents the neighbor list of a vertex, multiplying the row with matrix $X^{(l)}$ essentially `aggregate's features of neighboring vertices.
%In the perspective of a single vertex, it is the procedure of collecting the feature vectors of all neighboring vertices and aggregating into a single feature vector.
While summation is the most popular aggregation function, many aggregation functions other than summation is possible (e.g., max).
Nonetheless, the underlying computation pattern is identical to matrix multiplication.
The second, matrix multiplication between $\tilde{A}\cdot X^{(l)}$ and $W$ or $X^{(l)}$ and $W$, is called the \emph{combination} phase.
Because the elements within an entire row represents features belonging to a single vertex, the multiplication does not incur any information transfer between the vertices, but just `combine's the features contained within the vertex.
In other words, it does not require inter-vertex communication.
%The combination phase is local to each vertex, and does not invoke communication between different vertices. 
To each vertex, it can be thought as the feature vector being passed through a fully connected layer, where each vertex shares an identical weight matrix.

As analyzed in \cite{hygcn}, aggregation phase usually becomes the main bottleneck of the GCN execution.
Unlike the combination which is essentially a dense-dense matrix multiplication, the high sparsity of the graph topology (\A) and the low arithmetic density of the aggregation phase yields an extremely memory-intensive characteristics.
Together with the reason that it is hard to optimize, many work mainly focus on the aggregation phase. 
It is also worth noting that as depicted in \cref{eq:gcn}, the order between aggregation and combination does not affect the result of a GCN layer. 
However, depending on the shape of $W$, the width of the feature vector could increase or decrease, which can be used to reduce the execution time of aggregation~\cite{bidirectional}.

%\JL{anything more to say?}

\subsection{Baseline GCN Accelerator}
%\JL{This section was also reused in SGCN, but a little bit modified.}
\label{sec:acc}

\input{eval/moti_vsense}

\begin{figure*}
 \includegraphics[width=\textwidth, trim={-4mm 0 -2mm 0}, clip=true]{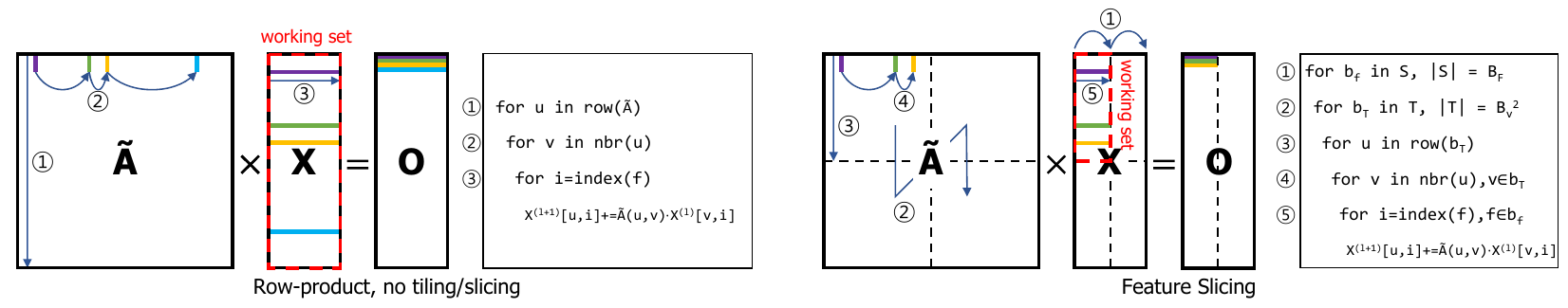}
%  \vspace{-3mm}
 \caption{Row product based GCN processing (left) and \fes (right).}
%\vspace{-3mm}
 \label{fig:rowproduct}
 
\end{figure*}

\begin{figure*}
 \includegraphics[width=\textwidth, trim={-4mm 0 -2mm 0}, clip=true]{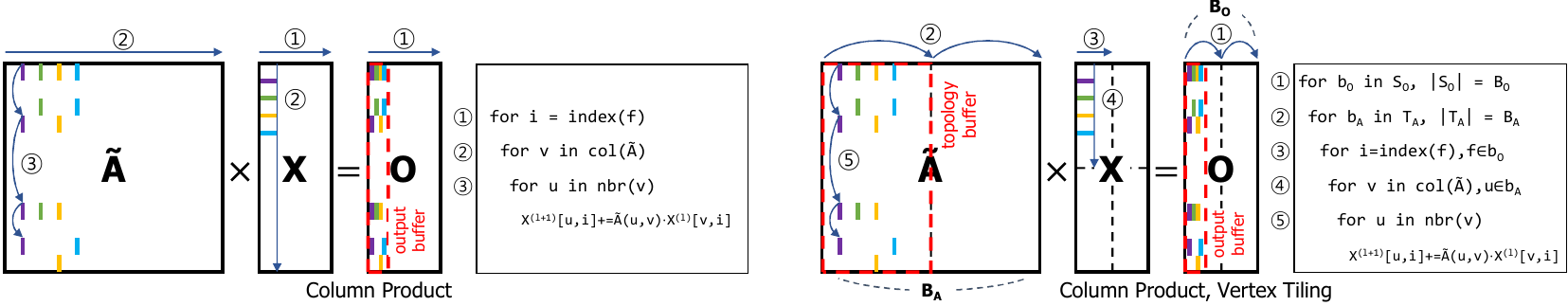}
%  \vspace{-3mm}
 \caption{Column product based execution (left) and its tiling (right).}
%\vspace{-3mm}
 \label{fig:colproduct}
 
\end{figure*}

\rev{
\figurename~\ref{fig:accelerator} presents a baseline GCN accelerator comprising aggregation engines and combination engines, similar to previous work~\cite{hygcn, dacgcn, engn, grip}, where the green component has been newly added that will be discussed in Section~\ref{sec:ctlr}.
The core of combination engine is a systolic array to support efficient matrix multiplication. 
Similar to previous systolic array based DNN accelerators~\cite{tpu, mbs}, there are two local memories in a double buffering manner that continuously provide input to the core.
The property buffer $X$, and the weight reader provide $W$, respectively. 
The output of the combination engine is read by the aggregation engine such that it can be processed in the aggregation function.

The aggregation phase is similar to classic graph processing, but each access to the edge contains a wide row of $X$ that has to be multiplied to the corresponding edge weight (i.e., $\tilde{A}_{i,j}$) and accumulated.
The SIMD cores process the multiply-accumulation. 
To feed the SIMD cores without stall, the two prefetch modules runs ahead in the graph topology such that the feature reader can always have on-the-fly reads to the feature matrix $X$.

The graph topology is stored in a compressed sparse row (CSR) format to exploit the sparsity of the data.
The vertex prefetch unit reads row pointers and passes it to edge prefetch and feature reader.
Then the edge prefetch unit reads column indices of CSR and sends them to the feature reader and SIMD cores.
When the feature reader receives an edge, it collects the rows of the feature matrix $X$. Because edges are sparse, the reads exhibit highly randomized pattern, making the aggregation phase memory-intensive and hard to optimize.

One way to mitigate the memory bottleneck in the aggregation phase is to use a large on-chip global cache. 
Shared by multiple engines in the accelerator, it temporarily stores the recently accessed features.
However, because of the wide feature matrix whose width often spans over hundreds to thousands, the working set usually well exceeds the size of the global cache, making it difficult to exploit localities.
}

\section{Motivational Example}
\label{sec:moti}
 
%\JL{I don't like this moti any more.. this shows what is bad, not what is good}

%\JL{should I move it next to the baseline?}

In GCN execution, the most commonly used loop tiling technique to reduce memory accesses is \emph{vertex tiling} as adopted from~\cite{gcnax, engn, awb}.
With vertex tiling, the adjacency matrix is split into multiple tiles and processed sequentially.
This way, the range of features being accessed is restricted within a tile to have a much smaller cache working set (and therefore exhibit more cache hits).
However, a better cache hit ratio incurs at the expense of multiple accesses on vertex features, either for reading or writing the features.
Because of this, choosing the right number of tiles becomes critical for high performance.

\figurename~\ref{fig:moti} illustrates sensitivity to the number of vertex tiles, according to graph datasets Pokec and Citation (See \tablename~\ref{tbl:dataset} for details.) with different feature width per vertex (32 and 128).
The Y axis shows speedup, normalized by the slowest among the number of tiles under consideration.

We make two observations from the plots: the performance is sensitive to the number of tiles, and the optimal number of vertex tiles differs according to dataset and feature width.
With the default 1$\times$1 vertex tiles, %(1$\times$1 at the baseline), 
much of the memory access count is spent reading the vertex features multiple times due to cache misses.
As the number of tiles increases, the working set size decreases and is better captured by the cache. %, leading to less input feature accesses. 
Simultaneously, the number of output feature access increases due to the repeated writes\footnote{In fact, it is possible to alter the order of tile traversal to increase the input access repetition instead. We omit this case as it mandates perfect caching.}. % caused by the computational pattern of the tiling (and not by the cache misses within tiles).
It is shown that the performance discrepancy between the best and the worst configuration greatly varies, up to 1.9$\times$ in the example (Pokec at $|F|=32$).
Furthermore, the relation between the amount of input access and number of tiles depends on the random access pattern posed by the adjacency matrix.
Not knowing the optimal setting a priori, one has to make an educated guess that can easily result in an inferior performance, or undergo several executions to find the optimal configuration which destroys the purpose of achieving speedup.

In this paper, we suggest \fes as an opportunity for on-line tiling configuration.
By exploiting the repetitive computation pattern found from \fes dataflow, 
the optimal configuration is gradually found over multiple rounds.
The proposed scheme not only searches for the optimal vertex tiling, but also the work distribution strategy which is crucial for multi-engine scaling of the GCN accelerators.

%address both issues, by placing the repetition cost on the topology data instead of \rev{the feature data} and deploying a self-tuning scheme.
%In this paper, by mitigating the trade-off between topology data and feature data access to reduce the total cost and deploying a self-tuning scheme, we address both issues.

% 

\section{\SaF}
\subsection{\FeS Dataflow}
\label{sec:dataflow}

Much of the prior work on GCN accelerations are based either on the \emph{row product}~\cite{hygcn, gcnax, dacgcn, engn} or \emph{column product}~\cite{awb}.
In this section, we describe the dataflow choice of \accname, which places \fes as the outermost loop that embraces vertex tiling.

\fig{fig:rowproduct} (left) displays the aggregation based on the basic row product dataflow (the innermost loop outputs a row). 
The colored strips of \A represent nonzeros in the topology matrix (i.e., edges) and the horizontal strips of $X$ are the elements of the feature vectors to be aggregated from the matching colored edges. 
In the row product execution model, \circled{1} the outermost loop visits the vertices in \A.
\circled{2} The next level loop visits the edges and \circled{3} aggregates the corresponding features. % using parallel processing units. 

The memory access pattern from this dataflow exhibits the following characteristics: %cost for this execution model considers three components. % into account.
First, reading \A is done sequentially and only once. 
%Because it is represented in a CSR format, its cost becomes $|V|+|E|$.
Second, features from neighboring vertices are read randomly over the entire feature matrix $X^{(l)}$, where a row of the feature matrix is read for each of the edges in \A.
Lastly, writing the output feature $X^{(l+1)}$ to the memory is also sequential and performed only once.
From the above characteristics, the performance bottleneck is at reading the feature matrix $X^{(l)}$.
With a global cache, the memory access can be reduced if the cache is large enough to capture the features, where the hit rate depends on the pattern of the edges.

When the feature matrix is significantly larger than the global cache size, the miss rate approaches 100\% and
loop tiling is a popular solution for mitigating this.
While there are many possible configurations, \fig{fig:rowproduct} (right) shows the choice of \accname named \emph{\fes}.
It partitions \A into $B_V \times B_V$ vertex tiles, and splits the feature matrix $X$ into multiple \emph{slices} of vertical strips.
As depicted in the figure, the two outermost loops (\circled{1} and \circled{2}) visit the feature slices and vertex tiles. 
The inner loops are identical to that from the \fig{fig:rowproduct} (left).
As observed in the dotted red lines, the working set is divided by the number of feature slices $B_F$ and number of tiles $B_V$, at the expense of reading the topology \A for $B_F$ times and reading the feature matrix $X$ for $B_V$ times. 

When $B_V$ and $B_F$ are large enough that all the tiles perfectly fit on the on-chip memory, only the cold misses remain.
In this manuscript, we call this strategy \emph{perfect tiling}.
While perfect tiling can maximize the cache efficiency, the required value of $B_V$ is usually too large.
Moreover, it neglects the sparsity of the graph topology and regards the computation as a dense matrix multiplication, often resulting in a suboptimal performance.

Column product is another dataflow solution for mitigating the cache miss. \fig{fig:colproduct} (left) shows the column product execution model. \circled{1} The outermost loop traverses each element along the feature vectors (i.e., the column of $X$). 
\circled{2} The next level loop visits the columns of \A. 
\circled{3} Lastly, the nonzero elements in the chosen column of \A are accessed, and the multiplied value is written to the corresponding position of the output matrix.

Because the two dimensions of $X$ are located at the outer loop, X is accessed only once.
However, it requires access to the adjacency matrix by the length of feature. 
For mitigating this, there is an way that use cache for adjacency matrix like \fig{fig:colproduct} (right).
When \A and $O$ are split into $B_A$ and $B_O$ vertical strips, respectively, a strip of \A stored in the cache is reused multiple times to improve performance.
but it needs an extra cache capacity and adds the cost of repeated access to the output matrix. 

In \accname, we place the feature slice visiting loop at the outermost level (\circled{1}). 
Even though the order of the two loops does not pose much difference in the working set size except for slight change in the pattern, 
it should be emphasized that this particular choice creates a recurring pattern over the random data access \circled{2}-\circled{5}.
This creates a unique opportunity for GCNs, where the inner loop configuration can be tuned over iterations of the loop \circled{1} because there is no difference among the feature slices, unlike the vertex tiles.
%If the vertex tile visiting loop \circled{2} is placed at the outermost level, each round would exhibit distinct access pattern due to the randomness of the graph data.
%This would make the optimal configuration from one round suboptimal to the next round.
Please note that this technique is infeasible for classic graph analytics, because the feature width is usually too narrow (e.g., 1 for PageRank).

The size of $B_F$ has to be decided such that the slice width $|F|/B_F$  must be at least over \SI{64}{\byte}, which is the granularity of the main memory and cache lines % (16 for 32-bit words)
to maintain the spatial locality. 
\accname sets the slice width to be 64 bytes (e.g., $B_F=16$ for $|F|=256$). 
%We will later show that this strategy is a rational design choice. \JL{in cost model and empirical results}

 \begin{figure}
 \centering
  \includegraphics[width=\columnwidth]{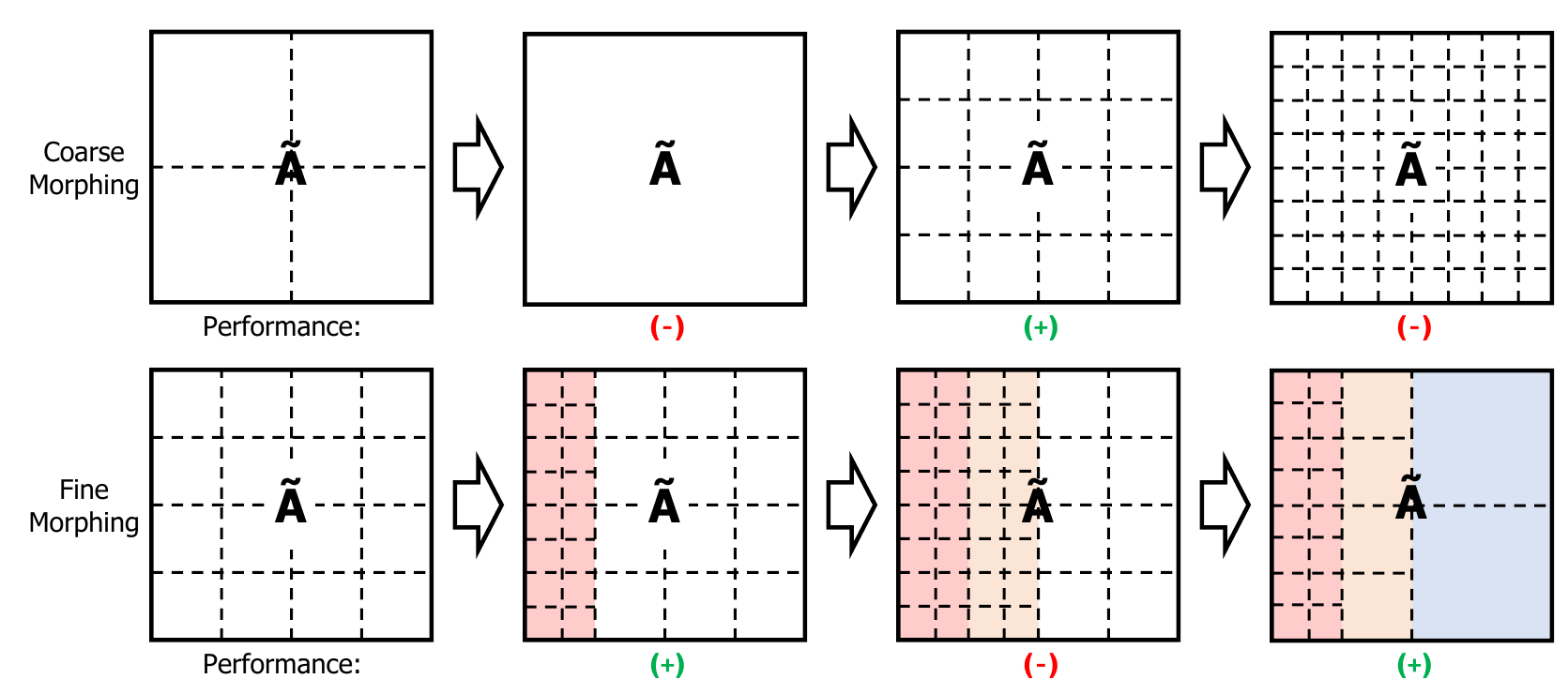}
 % \vspace{-3mm}
  \caption{Procedure of \atm.}
%  \vspace{-3mm}
  \label{fig:atm}
   \end{figure}
  
% \begin{figure}
% \centering
%  \includegraphics[width=.8\columnwidth]{figs/atm_eval2}
%  \caption{Example \atm over iterations.}
%  \label{fig:atm_eval}
%   \end{figure}

 \begin{figure*}
\centering
\includegraphics[width=\textwidth]{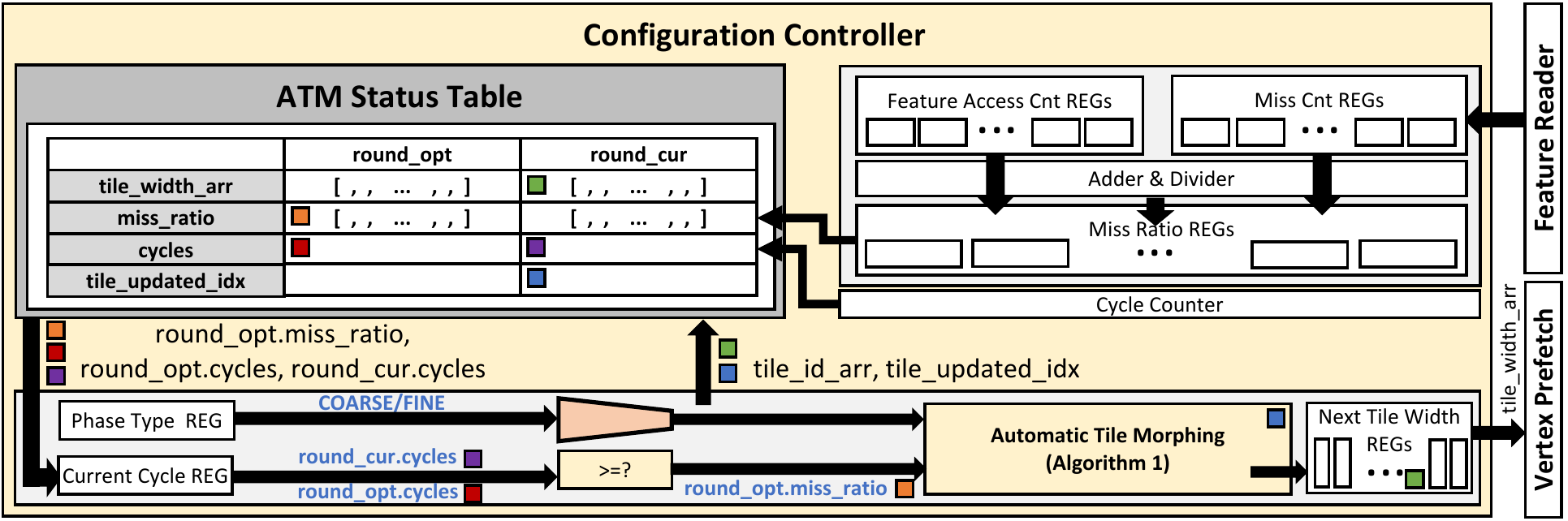}
\caption{Configuration controller architecture to support \atm.}  \label{fig:controller}
  \end{figure*}  

%\section{\SaF Design}

\subsection{\ATM}
\label{sec:atm}
As exemplified in Section~\ref{sec:moti}, choosing the right number of tiles is crucial to achieving high performance, yet there is no efficient method for optimal selection without an expensive trial-end-error procedure.
We propose \emph{\atm}, an approach that takes advantage of repetitive computational patterns to gradually configure the vertex tiling at runtime.

%\JL{Furthermore, the distribution of the topology is often heavily skewed~\cite{awb, leskovec}, and using uniform-sized tiles could result in suboptimal performance. -- can we strongly use this claim?}

With \fes dataflow (Section \ref{sec:dataflow}, the topology data \A is traversed multiple ($B_F$) times in the exact same pattern (\circled{2}-\circled{5} in \fig{fig:rowproduct} (right)). 
From each round\textemdash processing a slice of $X$\textemdash of the execution, \atm dynamically tunes the vertex tiling to reach the nearly optimal point at runtime, without the need for off-line design space exploration.
While a similar dynamic method can be devised by executing vertex tiling as outer loop and tuning $|F|$ as the parameter, the processing pattern of each tile would be distinct, making the tuning result from one tile unusable to the next.

   %\JL{check the term. round or iteration?}
   
   \fig{fig:atm} displays an example procedure of \atm, comprising two phases: \emph{coarse morphing} and \emph{fine morphing}.
   Coarse morphing (top) assumes that performance is a convex function of the number of tiles.
   Starting from a default setting (2$\times$2 in the figure), the performance is measured with twice more tiles in each dimension, and then half the number of tiles in each dimension. 
   In the direction of better performance, the same doubling or merging continues until a decrease in the performance is observed, the number of tiles is rolled back to that of the previous iteration (4$\times$4 in the example), and the coarse morphing terminates. 
   
Then, fine morphing (bottom) starts from the best partitioning of the coarse morphing phase.
The goal of fine morphing is to use heterogeneously sized tiles, such that the unbalanced graph distribution is considered.
Real-world graphs are sometimes collected by crawling, which is based on a breadth-first search order.
%Because of this, the regions with lower vertex IDs are likely to be denser than the other regions.
In addition, because of the small-world phenomenon~\cite{leskovec}, dense clusters often exist. 
These clusters exhibit a larger number of access instances and likely require a smaller working set compared to the sparser region of the graph.
We confined the tiling to have variable-sized vertical strips of square tiles to reduce the search space size, which effectively converts fine morphing to a one-dimensional partitioning problem.

Fine morphing records the hit rates of each tile and uses the rates to determine the visiting order. 
In the order of the lowest hit rate ({red} $\rightarrow$ {orange}), the vertical strips of blocks are further split into half-sized tiles, with one column per iteration. 
Similar to coarse morphing, this process is performed until performance degrades, and it rolls back the change that degraded the performance.
Then, similarly, but in the order of the highest hit rate (blue), two adjacent columns are merged until the performance decreases.
While we can further examine splitting a tile to 4$\times$4 tiles or merging 4$\times$4 tiles into one, we found no case in our dataset where such a configuration could further accelerate the execution. 
%\JL{can we provide a hitmap? or static analysis table?}

For implementation, we defined a unit tile size and stored the length of each vertical strip in the units. % of the smallest tile. 
In practice, the unit tile can be set as the tile size that perfectly fits the global cache.
In this paper, we split the topology into 64$\times$64 unit tiles.
In the worst case, there can be at most 64 unit-width strips, where storing them consumes less than a KB of capacity.

%   \rev{
% \fig{fig:atm_eval} shows how \atm performs over the repetitions in PK dataset.
% The first four iterations are coarse morphing.
% It gradually checks multiple $B_V$ values and settles at $B_V=8$.
% The overhead from these non-optimal executions is amortized over multiple iterations.
% In the fine morphing, it ends up further splitting the first column which has the lowest hit rate, and merging the last two columns which have the highest hit rate in our experiments, \atm was able to obtain over 95\% the performance that could be obtained by the optimal analysis, which is an unavailable information at runtime.
% }

%\revn{\fig{fig:atm_eval} shows how \atm performs over the repetitions in PK dataset.}
Empirically, \atm finds a solution near the static best.
To avoid the decision process from being on the critical path, we decide the next iteration's policy based on the measurements until the penultimate column of blocks.
However, a runtime overhead exists from the few first feature slices being executed with suboptimal tiling. % configurations. 
%When $B_F$ is large, the overhead
%Even though the execution times are longer than the optimal for a few iterations, these overheads 
%is amortized over the iterations. %, as we presented in Section \ref{sec:eval}.
When $B_F$ is too small, the portion of this overhead grows relatively large, which is another reason to favor a higher $B_F$.
Please see Section~\ref{sec:bf_sens} for further discussion.

%\JL{I removed the entire workload distribution. do i need it?}
\begin{comment}
%\section{Scaling the Accelerator}
%\subsection{Multi-Engine Accelerator Architecture}
\subsection{Workload Distribution}
A single set of a combination engine and aggregation engine can utilize more than 50\% of a single DDR4 channel (see Section~\ref{sec:scale}). 
However, we need to scale to multiple engines to design a high-end accelerator that feeds on more than 100 GB/s, similar to modern graphical processing units (GPUs)~\cite{gv100} or recent versions of NPUs~\cite{tpuv4i}. 
%Thus, we scale the accelerator by deploying multiples of each engine.
%While a simple scale-up (larger systolic array, wider SIMD) choice exists, it has been shown that those approaches often fall short due to the higher propagation delay or difficulty of finding enough parallelism from the workloads~\cite{deepstore, tpunas}.
%Similar problem exists, especially for aggregation engines. 
%If the SIMD way is larger than the width of the features, many of the SIMD core become idle.
%There is an option to process features from multiple neighbors at the same time, but it requires a redundant reduce phase.

One direct issue for deploying multiple engines is work distribution. %, which is directly connected to the performance. 
A naive strategy is equal vertex partitioning
%We set this as a baseline strategy 
as shown in \fig{fig:lac_moti} (a) performed over Pokec graph.
%

\end{comment}

%\input{eval/lac_moti}

\subsection{\SaF Architecture}
\label{sec:ctlr}

To support \atm, \accname adds a hardware component named \emph{configuration controller} to the accelerator shown in \figurename~\ref{fig:accelerator}.
The detailed architecture of the controller is illustrated in \figurename~\ref{fig:controller}. % along with Algorithm \ref{alg:alg1} and \ref{alg:alg2}.

The job of the configuration controller overall is to collect statistics given from the feature reader, make configuration decisions, and pass the decision to the vertex prefetch unit.
%Algorithm \ref{alg:alg1} and \ref{alg:alg2} describe how the controller decides next \atm and \lac policy.
%Feature slicing enables the inference to iterate total $B_F$ rounds of the same structure so that the controller runs $B_F$ times.
%The core aim of the controller is to find the best configuration that minimizes the miss ratio and cycle through multiple rounds.
%Controller manages three flags; \textit{atm\_type}, \textit{lc\_type}, \textit{phase\_type}.
For each iteration round of $B_F$, %controller calculates information and calls Algorithm \ref{alg:alg1} and \ref{alg:alg2}.
%Before Feature Reader sends a signal that this round is over, 
the controller accumulates the number of feature accesses and misses per vertical strip of unit tiles to calculate miss ratio. %computes miss ratio of each tile and gets running cycles of this round.
When the round is over, the controller updates these miss ratios and cycles into the \emph{ATM Status Table}. 
Based on this information from current round (round\_cur) and from the stored optimal (round\_opt), the controller invokes Algorithm \ref{alg:alg1} to produce next round's decision for the vertex tile size.
% The core of this algorithm is to find the best configuration that minimizes the miss ratio and cycle through multiple rounds.
% For each round, the controller collects accessed feature index and boolean value pairs which express each feature is hit or miss from the feature reader.
% Using this cache miss information, controller keeps accumulating misses which are used for determining policy until the feature reader sends a round end signal.
% After the round end signal arrived, the controller calculates miss ratios and stores these with running cycle of this round into ATM/LC Status Table (line 11-14).
%Refer to this table, Algorithm \ref{alg:alg1} and \ref{alg:alg2} decide \atm and \lac configuration and pass it to Vertex Prefetch.

\begin{figure*}
 \includegraphics[width=\textwidth, trim={-4mm 0 -2mm 0}, clip=true]{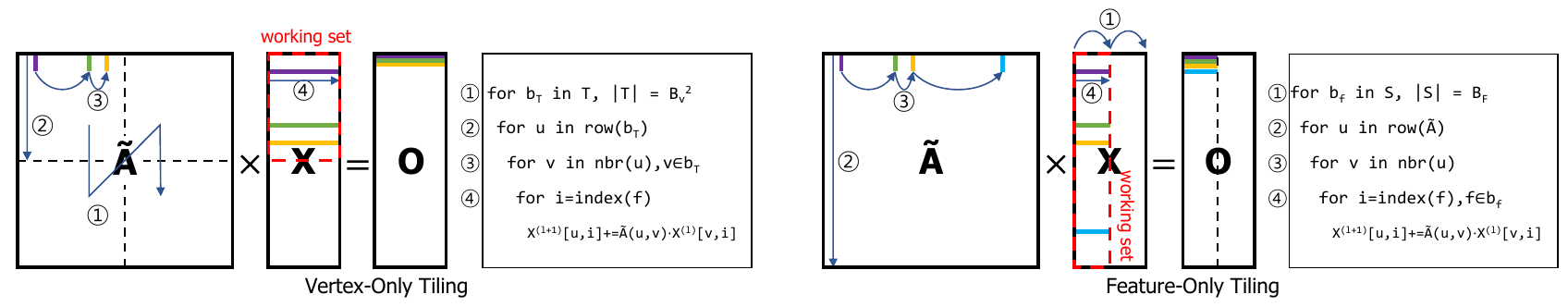}
%  \vspace{-5mm}
 \caption{Two extreme cases of loop tiling for vertex-only tiling (left) and feature-only tiling (right).}
\vspace{-3mm}
 \label{fig:extreme}
 
\end{figure*}

\begin{algorithm}
  \caption{Automatic Tile Morphing}
  \label{alg:alg1}
  \scriptsize
  \begin{flushleft}
 \hspace*{\algorithmicindent}\textbf{Input}:\\
 \hspace*{\algorithmicindent}\hspace*{\algorithmicindent}$round\_cur, round\_opt$: configuration of current/optimal round\\ 
 \hspace*{\algorithmicindent}\hspace*{\algorithmicindent}$atm\_direction$: current ATM morphing direction flag ($HALVING/MERGING$)\\
 \hspace*{\algorithmicindent}\hspace*{\algorithmicindent}$phase$: current ATM phase flag (\textbf{$COARSE/FINE$})\\ 
 \hspace*{\algorithmicindent} \textbf{Output}:\\
  \hspace*{\algorithmicindent}\hspace*{\algorithmicindent}$new\_tile\_width\_arr$: next tile width array decided by ATM\\
 \end{flushleft}
  \begin{algorithmic}[1]
  \scriptsize
    %\State {\color{blue(ncs)}{// If current round is optimal, update configuration}} 
    \If {$round\_cur.cycles < round\_opt.cycles$} {\color{blue(ncs)}{// if faster, update to optimal setting}}
        \State {$round\_opt \leftarrow round\_cur$}
    \ElsIf {$phase = COARSE$} {\color{blue(ncs)}{// if slower, enter fine morphing}} 
        \State {$phase \leftarrow FINE$}
    \Else \ {\color{blue(ncs)}{// for rest, this setting is optimal we found.}}
        \State{\Return {$round\_opt.tile\_width\_arr$}}
    \EndIf
    \If {$phase = COARSE$}
    % \If {$COARSE$ morphing phase}
        \If {$atm\_direction = HALVING$} {\color{blue(ncs)}{// increase the number of the tiles.}}
        % \If {we need to \textbf{halve} tiles}
            % \For {$i \leftarrow 0$ to $round\_cur.tile\_width\_arr$.size()}
            % \State{$tile\_width\_arr \leftarrow$ $2*$[$tile/2$ \textbf{for} $tile$ \textbf{in} $round\_cur.tile\_width\_arr$]} {\color{blue(ncs)}{(halving)}}
            % \For {$index$ in all tiles of $round\_cur.tile\_width\_arr$}
            \For {all tiles of $round\_cur.tile\_width\_arr$}
                %\State {\color{blue(ncs)}{// Halving: append $merged\_width/2$ twice}} 
                % \State{}{\color{blue(ncs)}{// twice}}
                % \State {$new\_tile\_width\_arr$.append$(round\_cur.tile\_width\_arr[index]/2)$}
                % \State {$new\_tile\_width\_arr$.append$(round\_cur.tile\_width\_arr[index]/2)$} 
                % \State {$tile\_width\_arr$.append$(round\_cur.tile\_width\_arr[i]/2)$}
                \State {append \textbf{halved} tiles to $new\_tile\_width\_arr$.} {\color{blue(ncs)}{// e.g. [4,4] $\rightarrow$ [2,2,2,2]}}
            \EndFor
            % \State{}
        \ElsIf {$atm\_direction = MERGING$} {\color{blue(ncs)}{// decrease the number of the tiles.}}
            % \State{$tile\_width\_arr \leftarrow$ [2*$tile$ \textbf{for} $tile$ \textbf{in} $round\_cur.tile\_width\_arr$[:$int(len(round\_cur.tile\_width\_arr)/2)$]]} {\color{blue(ncs)}{(merging)}}
        % \ElsIf {we need to \textbf{merge} tiles} {\color{blue(ncs)}{// we want to decrease the number of the tiles.}}
            % \For {$index$ in $0$ to $round\_cur.tile\_width\_arr$.size()$ / 2$}
            \For {half of tiles in $round\_cur.tile\_width\_arr$}
            % %     %\State {\color{blue(ncs)}{// Doubling: append $merged\_width*2$ once}} 
            %     \State {$new\_tile\_width\_arr$.append$(2 * round\_cur.tile\_width\_arr[index])$}
                \State {append \textbf{merged} tile to $new\_tile\_width\_arr$.} {\color{blue(ncs)}{// e.g. [2,2,2,2] $\rightarrow$ [4,4]}}
            \EndFor
        \EndIf
    \ElsIf {$phase = FINE$}
    % \ElsIf {$FINE$ morphing phase}
        \If {$atm\_direction = HALVING$} {\color{blue(ncs)}{// increase the number of the tiles.}}
        % \If {we need to \textbf{halve} tiles}
            \State {find \textbf{worst} miss ratio tile and \textbf{halve}.}
            % \State {$index \leftarrow$ index of \textbf{worst} miss ratio tile index}
            \State {append \textbf{halved} tiles and \textbf{other} tiles to $new\_tile\_width\_arr$.}
            % \State {append other tiles to $new\_tile\_width\_arr$.}
            % \State {\color{blue(ncs)}{// Find worst miss ratio tile $t\_id$ and Halve}} 
            % \State {$t\_id \leftarrow$ max\_index$(round\_cur.miss\_ratio)$}
            % \For {$i \leftarrow 0$ to $round\_cur.tile\_width\_arr$.size()}
            %     \If {$i = index$}
            %         \State{}{\color{blue(ncs)}{// twice (halving)}}
            %         \State {$new\_tile\_width\_arr$.append$(round\_cur.tile\_width\_arr[i]/2)$}
            %         \State {$new\_tile\_width\_arr$.append$(round\_cur.tile\_width\_arr[i]/2)$}
            %     \Else
            %         \State{}{\color{blue(ncs)}{// maintain other tile setting}}
            %         \State{$new\_tile\_width\_arr$.append$(round\_cur.tile\_width\_arr[i])$}
            %     \EndIf
            % \EndFor
        \ElsIf {$atm\_direction = MERGING$}  {\color{blue(ncs)}{// decrease the number of the tiles.}}
        % \ElsIf {we need to \textbf{merge} tiles}
            % \State{}{\color{blue(ncs)}{// similar with halving.}}
            \State {find \textbf{best} miss ratio tile and \textbf{merge} with adjacent tile.}
            \State {append \textbf{merged} tile and \textbf{other} tiles to $new\_tile\_width\_arr$.}            
            % \State {\color{blue(ncs)}{// Find best miss ratio tile $t\_id$ and merge}}
            % \State {$t\_id \leftarrow $min\_index$(round\_cur.miss\_ratio)$}
            % \For {$i \leftarrow 0$ to $tile\_width\_arr$.size()}
            %     \If {$i = t\_id$} 
            %         \State {\color{blue(ncs)}{// Merge with adjacent tiles}} 
            %         \State {$tile\_width\_arr$.append$(round\_cur.tile\_width\_arr[i]$} 
            %         \State {$\quad\quad\quad\quad\quad\quad\quad\quad\quad$ + $round\_cur.tile\_width\_arr[i+1])$}
            %     \ElsIf {$i = t\_id + 1$}
            %         \  {\textbf{continue}}
            %     \Else
            %         %\State {\color{blue(ncs)}{// Use current $merged\_width$}}
            %         \  {$tile\_width\_arr$.append$(round\_cur.tile\_width\_arr[i])$}
            %     \EndIf
            % \EndFor
        \EndIf
        % \State {give updated tile index to $round\_cur.tile\_updated\_idx$}
        \State {$round\_cur.tile\_updated\_idx \leftarrow index$}
    \EndIf
 %   \State {\color{blue(ncs)}{// Save result and return}} 
    % \State {Update tile information to $round\_cur.tile\_width\_arr$.}
    \State {$round\_cur.tile\_width\_arr \leftarrow new\_tile\_width\_arr$}
    \State \Return {$new\_tile\_width\_arr$}
  \end{algorithmic}
\end{algorithm}

%Controller manages three flags; \textit{atm\_type}, \textit{lc\_type}, \textit{phase\_type} for deciding polices.
%By Feature Slicing, the inference iterates total $B_F$ rounds(line 5).
%Every inference round, controller collects accessed feature index and boolean value that expresses each feature is hit or miss from the Feature Reader.  
%Also, Feature Reader gives whether the round is end. 
%From the information, controller finds which tile is accessed and calculates miss ratio until this round ends (line 6-10).
%After the round overs, controller stores miss ratio and running cycle to ATM/LC Status Table(line 11-14).
%Finally, controller decides \atm and \Lac policy and pass the configuration information to Vertex Prefetcher.

Algorithm \ref{alg:alg1} shows the simplified procedure of \atm.
The controller manages tiling by $\mathit{tile\_width\_arr}$, an array that stores how many unit tiles comprise each vertical strip. %contains the information about how many (width) each tile merged with adjacent tiles.
For example, it stores 32 consecutive 2's to represent 32\texttimes 32 tiles and 16 consecutive 4's to represent 16\texttimes 16 tiles. 
%At first, controller regard adjacency matrix as tiled into default size (e.g. 32) of unit tiles.
%In this configuration, $tile\_width\_arr$ is 32 number of 1s, because unit tiles are not merged with another tiles yet.
%If each adjacent unit tiles are merged, so that total tile size is 16, $tile\_width\_arr$ is 16 number of 2s, because each tile is consists of 2 unit tiles.
\Atm starts with updating the optimal value ($round\_opt$) in the status table (line 2).
%Based on the analysis on how cycle trend changes by round goes, 
The controller proceeds into the \textit{coarse morphing} phase.
Assuming that the performance-improving direction $\mathit{atm\_direction}$ have been found in the previous rounds to be halving the tile sizes (omitted in the algorithm), 
the new $\mathit{tile\_width\_arr}$ is populated with twice the original elements from the array of the current rounds of half the values (line 8-10). 
The merging direction functions in a similar manner (line 11-13). %JL: until here
%$phase$ and $atm\_direction$ (halve tile or merge tile) of \atm.
%Detailed explanation about this analysis is omitted because \atm process is simple convex problem and to show algorithm concisely.
In \textit{fine morphing} phase, controller finds the target tile ($t\_id$) that should be merged with adjacent tile or halved based on the miss-ratio (line 14-21).
%Tiles that showed high miss-ratio should be halved and tiles that showed low miss-ratio gets chance to be merged with adjacent tile to minimize total cycle.
After $\mathit{tile\_width\_arr}$ is determined, this tiling information is passed to vertex prefetch unit.
%Additionally, if fine mophing phase updated tiling configuration, updated tile index is returned, so that Algorithm \ref{alg:alg2} can use it.figuration, updated tile index is returned, so that Algorithm \ref{alg:alg2} can use it.

\JS{
When doing coarse morphing phase cannot make any performance gain, the controller starts \textit{fine morphing phase}.
The fine morphing phase fine-tunes tiling configuration based on the coarse tiling configuration found in the coarse phase.
In fine morphing phase, controller halves the target tile or merges the target tile with its neighborhood tile.
The target tiles are picked based on its hit-ratio which means }

The configuration controller stores access/miss/ratio registers for each of the 64 unit vertical columns.  %3*4B*64 = 768 B
In addition, $\mathit{tile\_width\_arr}$ and $miss\_ratio$ of the status table are 64 entry-long arrays for each of $\mathit{round\_op}t$, $\mathit{round\_cur}$, and at the output to the vertex prefetch. % 64 * 4B * 3 = 768 B. 
Thus, using 4 byte elements, the total size of the registers and status table is only about 1.5KB, a negligible size compared to the global cache. %\JL{chk this} 

\subsection{Dataflow Cost Comparison}
The \fes dataflow adopted in \accname implicitly prioritizes increasing the number of slices $B_F$ over increasing the vertex tiles $B_V$.
While it is true that this choice somewhat puts a restriction on loop tiling, we analyze the cost model to show that prioritizing is a advantageous strategy in general.

Using the terms Section~\ref{sec:gcnbackground} and $|E|$ as the number of edges, the memory access cost from the row product execution in \figurename~\ref{fig:rowproduct} (left) can be modeled as below.
\begin{equation}
%\begin{medsize}
\begin{aligned}
    M_{Row} &= (|V| + |E|) + m(|V||F|)\cdot|E||F| + |V||F|.
\end{aligned}
\label{eq:base}
%\end{medsize}
\end{equation}
The first and third terms account for sequentially reading the topology data in CSR format and writing the output matrix, respectively.
The second term, on the other hand, models the random reads to the feature matrix.
For each edge ($|E|$) in the topology, a row of the feature matrix is read ($|F|$).
Under the assumption that the miss rate $m(x)$ is a function of the working set size $x$, 
the second term represents the accesses filtered by the global cache.

Applying loop tiling, the bottleneck at the second term is reduced by reducing the working set size.
Using the tiling from \figurename~\ref{fig:rowproduct} (left), the model becomes:
\begin{equation}
%\begin{medsize}
\begin{aligned}
     \hspace{-2.5mm}M_{FS} & = B_F(B_V |V| + |E|) + m(\frac{|V||F|}{B_F B_V})\cdot |E||F| + B_V |V||F|.
\end{aligned}
    \label{eq:vt_fs}
%\end{medsize}
\end{equation}
The working set size is divided by the product of $|B_F|$ and $|B_V|$, where those factor are multiplied to reading the topology data and writing the output matrix, accounting for the repetitions.

Thus, intuitively, increasing $|B_F|$ yields a lower cost if the topology data are smaller, and increasing $|B_V|$ is better if the feature matrix size is smaller.
Because the feature width in GCNs usually reach several hundreds, it is often beneficial to favor increasing $|B_F|$ over $|B_V|$.
Consider two extreme cases where we only tile the topology ($|B_F|=1, |B_V|=B$) as in \figurename~\ref{fig:extreme} (left), versus only tiling the feature matrix ($|B_F|=B, |B_V|=1$) as in \figurename~\ref{fig:extreme} (right), such that the working set sizes are equal.
Substituting the values into 
Eq.~\ref{eq:vt_fs} and finding the condition for the former being smaller than the latter, 
\begin{equation}
%\begin{medsize}
\begin{aligned}
B|V|+|E|+B|V||F| &> B(|V|+|E|)+|V||F|, \\
    |F| &> |E|/|V|. 
\end{aligned}
    \label{eq:compare}
%\end{medsize}
\end{equation}
In the above equation, $|E|/|V|$ is the average degree of the graph, one of the representative parameters of graph data.
It is widely known that the average degree of real-world graphs is usually only around 10-20~\cite{ogb}.
Comparing this to $|F|$ being several hundreds, favoring increasing $|B_F|$ first can be a rational choice.

\section{Multi-Chip Scaling}
\label{sec:mcm}

\begin{figure}
    \centering
    \includegraphics[width=\columnwidth]{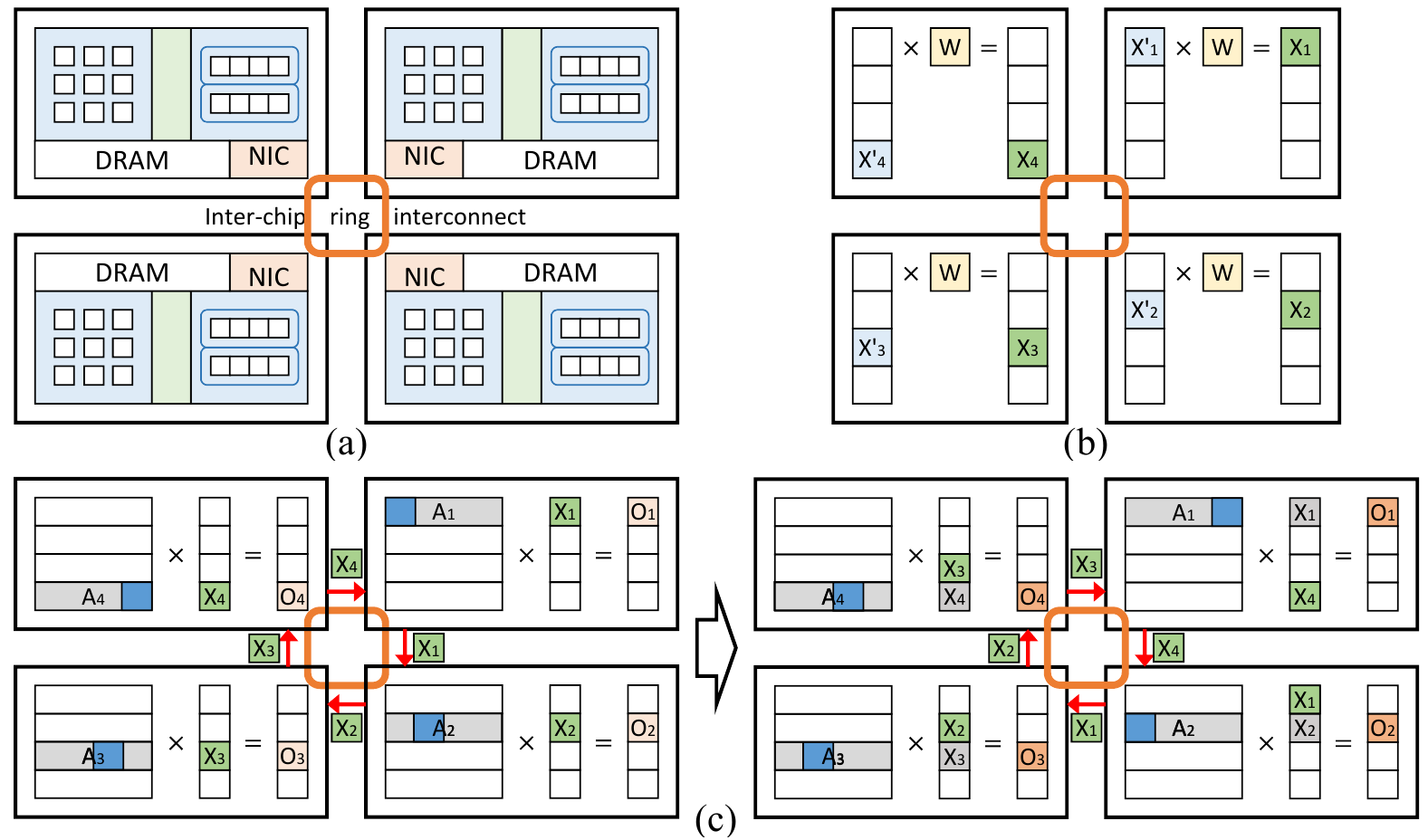}
    % \vspace{-1mm}
    \caption{Multi-chip scaling strategy.}
    % \vspace{-2mm}
    \label{fig:multichip_arch}
\end{figure}

\figurename~\ref{fig:multichip_arch} (a) illustrates the multi-chip environment. 
In multi-chip scaling, we assume that each chip is packaged with its own memory module such as HBM.
With such an environment, the scaling challenge is that communications need to be explicit because the memories are not shared.
\figurename~\ref{fig:multichip_arch} (b) and (c) show the strategy for scaling.
For the combination (b), we duplicate $W$ in the memory of each chip. 
This is feasible since $W$ is small. 
Then $X$ is split horizontally, such that each chip can produce its own piece of the output matrix.

For aggregation, the topology $\tilde{A}$ is similarly split into four pieces, and does not need to be communicated between the chips. 
However, $X$ needs to be communicated because each chip only has a part of $X$ but eventually needs the whole matrix.
In our strategy as in \figurename~\ref{fig:multichip_arch} (c) similar to all-gather~\cite{mpich}, each chip starts by aggregating only with the current piece of $X$. 
At the same time, the same piece is passed to the neighboring chip in the ring.
Next, the data that has been just received piece is being processed and passed to the neighbor.
After $N-1$ steps for $N$ chips, the aggregation is complete.
%
%we adopt the ring-based all-gather algorithm~\cite{mpich} to perform the communication.
Note that there is no further communication after aggregation (Except for the last GCN layer), because the split output will be directly used as split input to the next layer's combination.

\section{Evaluation}
\label{sec:eval}

\subsection{Dataset}

\begin{table}[t]
\footnotesize
\centering
\caption{Benchmark Dataset Information}
\begin{tabular}{ccccccc}
\toprule
Dataset & \#Vertices & \#Edges  & \#Features & Topology & Feature \\ 
\midrule
%    \cmidrule(lr){1-1}
%    \cmidrule(lr){2-5}
Products (PD) &2.45 M&61.9 M&100 &  0.24 GB  &   0.91 GB \\
Citation (CT) &2.93 M&30.6 M&128 & 0.12 GB  & 2.79 GB \\
Pokec (PK) &1.63 M&30.6 M&256 & 0.12 GB  &  1.56 GB \\
YouTube (YT) &1.13 M&2.99 M&256&0.02 GB  &  1.08 GB \\ %com-YouTube
%WikiTalk (WT) &2.39 M&5.02 M&256& 0.03 GB  &   2.28 GB \\ %wiki-topcats
LiveJournal (LJ) &4.85 M&69.0 M&256 & 0.28 GB  &   4.63 GB \\
Orkut (OK) &3.07 M&117 M&256 & 0.45 GB  &  2.93 GB \\
Reddit (RD) & 0.23 M & 114 M & 602 & 0.43 GB  &  0.52 GB \\
 \bottomrule
\end{tabular}
\label{tbl:dataset}
\end{table}

We use seven large graph datasets from the real world as benchmarks.
The details are listed in \cref{tbl:dataset}.
Products (\emph{PD}) is from Open Graph Benchmark Collection~\cite{ogb}.
It has 2.45 M vertices that represent Amazon products with 61.9 M edges which indicates products being purchased together.
Each vertex has word embeddings of width 100 as features, generated from product descriptions. 
Citations (\emph{CT}) is also from the Open Graph Benchmark Collection and represent 2.93M papers as vertices and their citations as 30.6M edges from Microsoft Academic Graph~\cite{mag}.
Each vertex is associated with word embeddings from their titles and abstracts.
Pokec (\emph{PK}), YouTube (\emph{YT}), LiveJournal (\emph{LJ}), Orkut (\emph{OK}) and Reddit (\emph{RD}) are from Stanford Large Network Dataset Collection~\cite{snap}.
Pokec~\cite{pokec} is a social network graph from Slovakia, and includes 1.63M users as vertices and 30.6M connections as edges.
Youtube~\cite{youtube} is graph from users of a video sharing service, with 1.13M users and 2.99M connections between them.
Livejournal~\cite{livejournal} and Orkut are from online social network services with 4.85M and 3.07M vertices, respectively.
Reddit~\cite{reddit} is a social network service, and 0.23M vertices are collected from the Reddit discussion forum.
The features of the Reddit dataset consist of 302-dimensional post information, and GloVe CommonCrawl~\cite{pennington2014glove} based 300-dimensional word embedding vectors.
For others, the node embedding vectors of width 256 have been generated from node2vec~\cite{node2vec} and used as features.

\subsection{Methodology}

\begin{table}[t]

%\scriptsize
\footnotesize
\centering

\caption{System Configuration}
\begin{tabular}{ccccc}
\toprule
%\multicolumn{3}{c}{\textbf {Common}}\\
\multirow{6}{*}{\textbf {Common}} & \multirow{3}{*}{\makecell{Accelerator\\Engine}} & Frequency & 1GHz \\
&& Combination & 32$\times$32 Syst. Array \\
&& Aggregation & 16-Way SIMD \\
\cmidrule(lr){3-4}
&\multirow{3}{*}{Global Cache} & Capacity & 16MB\\
&& Ways & 16 \\
&& Replacement & LRU \\
\midrule

\multirow{6}{*}{\makecell{\textbf{Multi-engine}}}  & \multirow{2}{*}{\#Engines} & Aggregation & 8 \\
&& Combination & 8 \\
\cmidrule(lr){3-4}
&\multirow{4}{*}{\makecell{Off-chip\\ Memory \\(Shared)}} & Spec. & HBM2 \\
&& Peak Bandwidth & 256 GB/s \\
&& Channels & 8 \\
&& Banks & 4$\times$4 \\
\midrule

\multirow{6}{*}{\makecell{\textbf{Small scale}}} & \multirow{2}{*}{\#Engines} & Aggregation & 1 \\
&& Combination & 1 \\
\cmidrule(lr){3-4}
&\multirow{4}{*}{\makecell{Off-chip\\ Memory}} & Spec. & DDR4-2666 \\
&& Peak Bandwidth & 21.3 GB/s \\
&& Ranks & 4 \\
&& Banks & 4$\times$4 \\

\midrule

\multirow{5}{*}{\textbf {Multi-Chip Module}} & \multirow{2}{*}{\#Modules} & Accelerator & 1-32 \\
&& HBM & 1-32 \\
\cmidrule(lr){3-4}
&\multirow{3}{*}{\makecell{Interconnect}} & Topology & Ring \\
&& Bandwidth & 256 GB/s \\
&& Latency & \SI{20}{\nano\second} \\

 \bottomrule
\end{tabular}
\label{tbl:system}
% \vspace{-1mm}
\end{table} 

\begin{table}[t]
%\begin{tcolorbox}[colframe=olivegreen,colback=white,left=0pt, right=0pt]
\scriptsize
\centering
\caption{Comparison of Existing GCN Accelerators}
\setlength\tabcolsep{3pt}
    \resizebox{\columnwidth}{!}
    {
\begin{tabular}{cccccc}
\toprule
\multirow{3}{*}{Scheme} & \multirow{3}{*}{Dataflow} & \multicolumn{2}{c}{Loop Tiling} & \multicolumn{2}{c}{Tiling Configuration}  \\
\cmidrule(lr){3-4}\cmidrule(lr){5-6}
&& Topology & Feature  & Method & \makecell{Non-perfect\\Tiling} \\
\midrule
HyGCN & Row Product & \xmark & \xmark & \xmark & N/A \\
EnGN &  Row Product & \xmark & \cmark  & \xmark & N/A\\
AWB-GCN & Col. Product & \cmark & \cmark & \xmark & N/A \\
GCNAX & Row Product & \cmark & \cmark & Preproc. & \xmark \\
\midrule
\makecell{\accname \\(Proposed)} & Row Product & \cmark & \cmark & \makecell{ATM\\(Runtime)} & \cmark  \\
 \bottomrule
\end{tabular}
    } % resizebox
\label{tbl:schemes}
%\end{tcolorbox}
\vspace{-1mm}
\end{table}

\rev{
We evaluate the performance of \accname architecture with the configuration listed in \cref{tbl:system}.
The accelerator runs in 1GHz frequency, where the combination engine has 32\texttimes32 systolic array and the aggregation engine has 16-way SIMD cores.
Both engines use 32bit fixed point arithmetic operations.

We use HyGCN as baseline architecture, as HyGCN has an easily expandable architecture for vertex tiling and feature slicing.
On the other hand, GCNAX uses static vertex tiling by pre-profiling the workload and AWB-GCN does not use caches for exploiting locality. 
Therefore, applying \accname which uses dynamic strategy and cache on such architectures would be infeasible.

We conduct the experiments in total of three different settings.
In our default setting, a multi-engine configuration is used, which comprises eight of both engines, with an HBM2 memory subsystem. %to provide high bandwidth from the memory.
In our small scale setting, we show that \accname is also efficient on a design with small resource budget and low memory bandwidth.
The setting includes one of both engines with a 4-rank DDR4 channel.
Lastly, we demonstrate the extension to the multi-chip scaling method described in \cref{sec:mcm}. 
We use one to 32 chips connected in a ring topology network.
Following other MCM accelerators~\cite{simba,mcmgpu} we use 256GB/s bandwidth ring interconnect, with \SI{20}{\nano\second} per-hop latency, where each chip is connected to its own HBM module. 
To validate the design, we implemented the accelerator in Verilog HDL and synthesized it using Cadence Genus with 45~\SI{}{\nano\meter} OpenPDK library, which was scaled to 32~\SI{}{\nano\meter}.
To model the on-chip memory, we used CACTI~\cite{cacti} to draw the power and area under \SI{32}{\nano\meter} technology node.
The baseline chip resembling HyGCN~\cite{hygcn} consumed 11.40~\SI{}{\square\milli\meter}, 
and \accname has an area of 11.55~\SI{}{\square\milli\meter}.
For comparison we reproduced GCNAX~\cite{gcnax} and AWB-GCN~\cite{awb}, which consumed 11.23 \SI{}{\milli\meter\squared} and 13.80 \SI{}{\milli\meter\squared}, respectively.

For performance modeling, we ported SCALE-Sim~\cite{scalesim} to C++, and extended it to support the \accname's aggregation unit.
The cycles were matched to that of the RTL model.
For modeling the DRAM subsystem, we used DRAMsim3~\cite{dramsim3}.
We faithfully reproduced the previous work from \cref{tbl:schemes}.
HyGCN~\cite{hygcn} is the baseline accelerator with hybrid engines, and EnGN~\cite{engn} proposes an RER dataflow with DAVC that captures high-degree vertices.
AWB-GCN~\cite{awb} is based on column product which utilizes same set of PEs for both aggregation and combination with separate task distributors.
Lastly, GCNAX~\cite{gcnax} perform a static analysis to find the optimal loop tiling under the assumption of perfect tiling.
For comparison, we also added a ‘VT’ version of the accelerator, to which we normalized all values.
}

% \rev{
% Comparing VT and \accnameplus\textsubscript{Optimal} provides an analysis on efficiency of our proposed feature slicing technique and helps choosing the right partitioning method for GCN inference.
% }
% \Fes outperforms VT, by up to 20.0\% speedup. 
%It is worth noting that 
%VT outperforms \accnameplus\textsubscript{Optimal} only on RD dataset.
%In this case, due to the $B_F$ size limitation, the ideal working set size requires a $B_F$ value higher than the limit.  

\subsection{Multi-Engine Accelerator Results}
\label{sec:scale}

\input{eval/single_perf_side}

\rev{
In this subsection, we analyze the multi-engine performance to show the proposed \accname outperforms the existing accelerators.
We placed eight engines into the accelerator, with HBM2 memory providing 256 GB/s peak bandwidth. 
The results are shown in \figurename~\ref{fig:multi_perf}.
We set `VT' as a baseline which only performs vertex tiling. 
‘\accname’ is the proposed accelerator with the \fes on top of vertex tiling with \atm. 
Among the multiple engines, we split the graph workload such that the number of edges per engine is equal.
For all implementations except \accname which is dynamic, we examined all possible partitioning combinations off-line (i.e., optimal) and chose the best one.

Comparing \accname with VT gives an insight into how much benefit can be drawn from choosing the right loop tiling scheme for GCN inference.
The performance improvement of \accname over VT is 73.1\%, obtained by reducing the significant amount overhead of VT's repetitive feature access cost and dynamically finding the right configuration.
The largest speedup is on PD with 2.76\texttimes, coming from the fact that the optimal setting of VT was at a large $B_V=64$.
This large $B_V$ makes a huge repetition overhead to the output feature accesses.
\Fes can efficiently split these repetition counts, contributing to the superior speedup. %by reducing the repetition count ($B_V$) from 64 to 16.

The proposed technique outperforms loop tiling strategies from the existing prior designs in all datasets.
HyGCN adopts no tiling, and shows a poor cache efficiency.
EnGN is based on vertex tiling, and uses DAVC to capture features from high-degree vertices. 
However, we have found that the DAVC has a marginal effect on the speedup.
One reason is that there are too many high-degree vertices in the large-scale real-world graph datasets for DAVC to capture.
Moreover, when their reuse distances are short enough, they are often already well-captured by the default LRU replacement policy.
AWB-GCN is unique among the designs under test in that they utilize column product based execution with an aggressive load balancing.
It requires a non-negligibly large output buffer to function without a separate consideration for spilling, given that there are a few million vertices in the graph.
Even though AWB-GCN splits the matrix, it adds the repetition counts on the feature accesses, which becomes a reason to underperform \accname by 35.7\%. 
GCNAX provides an analytic model for finding the optimal partitioning scheme, and it provides 15.4\% speedup over VT.
However, they only consider perfect tiling in their model, missing an important opportunity for performance enhancements.
Moreover, because they unrealistically assume uniformly distributed sparse graphs, their model is often distant from the real measurements and this becomes another source of 68.2\% performance loss compared to \accname.

}

\begin{figure}
\centering
 \includegraphics[width=0.9\columnwidth]{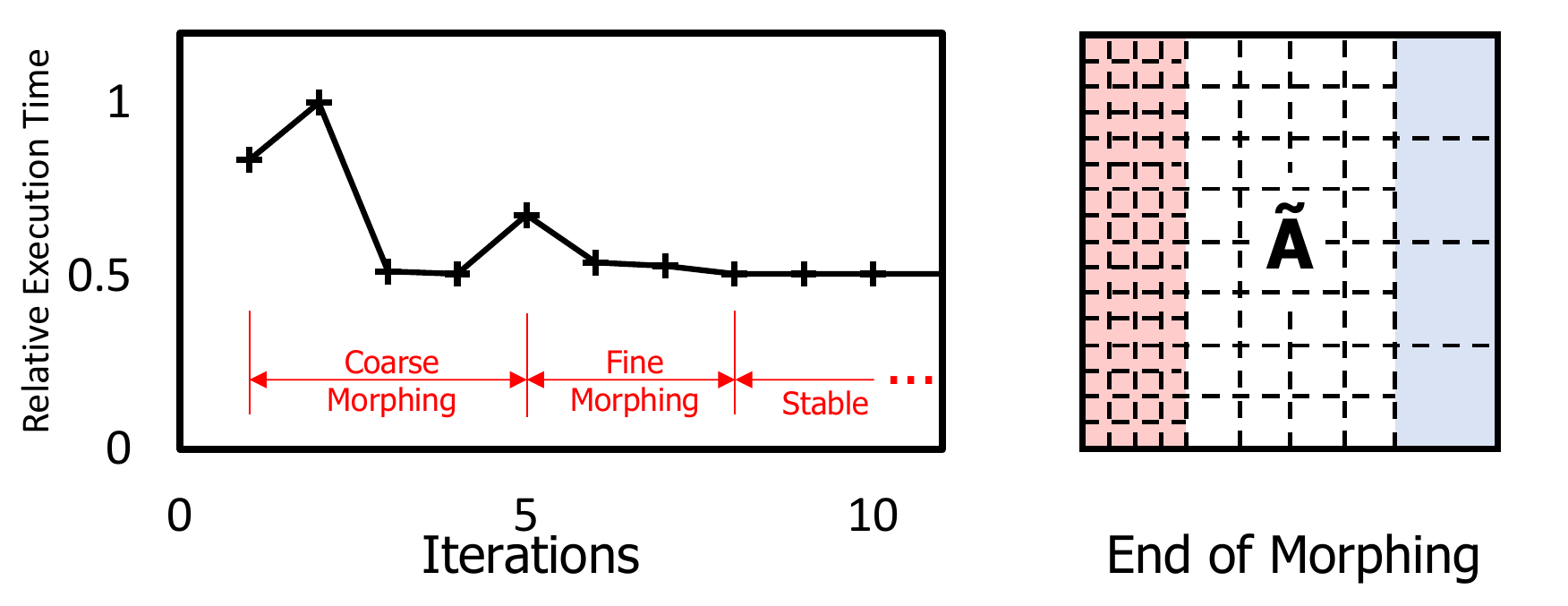}
 \caption{Example \atm over iterations.}
 \vspace{-3mm}
 \label{fig:atm_eval}
  \end{figure}

  \begin{comment}
  \accnameplus\textsubscript{Dynamic} shows how the \atm quickly finds the near-optimal performance without off-line analysis.
In geometric mean, it achieves 96.3\% of the performance of the optimal settings found by static analysis (\accnameplus\textsubscript{Optimal}).
Note that \accnameplus\textsubscript{Dynamic} is still significantly better than the prior designs which require static analysis.
%The gap is relatively larger for datasets that have large difference between \accnameplus\textsubscript{Optimal} and \accnameplus\textsubscript{Dynamic} because they require more repetitions to converge to the right configuration. 
  \end{comment}
  
  \rev{
\fig{fig:atm_eval} shows how \atm performs over the repetitions in PK dataset.
The first four iterations are coarse morphing.
It gradually checks multiple $B_V$ values and settles at $B_V=8$.
The overhead from these non-optimal executions is amortized over multiple iterations.
In the fine morphing, it ends up further splitting the first column which has the lowest hit rate, and merging the last two columns which have the highest hit rate in our experiments, \atm was able to obtain over 95\% the performance that could be obtained by the optimal analysis, which is an unavailable information at runtime.
}

% It is worth noting that the FS only version of \accname resulted in a performance degradation. The reason comes from the fact that RD has a large average degree, of almost 500. 
% This aligns with our cost model in Eq.~\ref{eq:fsvt}, because the number of features has to be larger than the average degree, which is not the case in reddit.

\subsection{Smal-Scale Accelerator Results}
\rev{In this subsection, we analyze the single-engine accelerator performance to show that \fes and \atm are also effective for systems under tight resource budget such as mobile and embedded systems.
\figurename~\ref{fig:single_perf} shows the small scale performance of the GCN execution.

}

As in multi-engine settings, \accname outperforms all prior designs.
One interesting aspect is that among the prior designs GCNAX performs better than AWB-GCN in the small scale setting.
In this setting, the available memory bandwidth is quite small, and it becomes crucial to utilize the cache efficiently.
Because AWB-GCN does not have particular caching mechanism, the benefit of better utilizing the cache becomes more important than load balancing.
\accname still finds better settings on all datasets except for OK and RD, with 10.8\% improvement over GCNAX and 46.4\% over VT.

Reddit shows a unique trend for performance speedup because it has the largest number of edges per vertex (dense topology) among the datasets while having fewer vertices.
The dense topology of RD works well with VT, as illustrated in~\figurename~\ref{fig:perf}, in which VT shows superior speedup over HyGCN.
Also, fewer vertices require a smaller feature matrix than other datasets, which diminishes the output write overhead of VT.
GCNAX uses perfect tiling, so it constructs a too large $B_V$ tiling strategy, which makes lower speedup over VT.
For \accname, the dynamic tiling found by the automatic tile morphing is almost equal to that of the static solution found by GCNAX.
Because automatic tile morphing takes some iterations to settle around the best solution, a slight speedup degradation of 0.82\% over GCNAX occurs for \accname as depicted in~\figurename~\ref{fig:single_perf}.
OK is also as dense as Reddit, but it has many vertices. 
This environment is not friendly for VT.
So, both GCNAX and \accname outperform VT, unlike the Reddit case.
\accname has a minor speedup degradation of 1.8\% over GCNAX, similar to Reddit.

% The OK dataset had the largest number of edge per vertex among the datasets.
% As a result, the dynamic tiling found by the \atm was almost equal to that of the static solution found by GCNAX.
% Because \atm took some iterations to settle around the best solution, a small degradation of 1.8\% was observed for \accname.

\subsection{Energy Consumption}
\input{eval/single_E}

\fig{fig:single_E} shows the normalized energy consumption of the aggregation phase with the breakdown, under the single chip, multi-engine setting.
We set the baseline as the architecture with no partitioning, whose energy consumption is depicted with shaded bars. %\NK{shaded? striped?}
The energy consumption of the \accname is represented by the solid bars.
As indicated in the figure, \accname consumes slightly more energy on computation, but much more energy is saved from the memory and the cache.
These both show the efficient usage of cache in \accname due to the increased cache hit rate,
% These both come from the increased cache hit rate, 
because less data is read from the DRAM, and less data is written to the cache.
An exception is LJ, where the benefit from cache and DRAM is slightly less than the increased energy in computation.
We can save 53\% energy on RD because it has an extremely dense topology structure that accompanies a smaller topology access ratio over feature accesses.
Therefore, since \accname can significantly reduce feature accesses, the energy consumption is greatly reduced.

\input{eval/igcn_perf}

\subsection{Effect of Reordering} 
Another promising scheme for enhancing cache efficiency for GCNs is graph reordering as adopted in Rubik~\cite{rubik} and I-GCN~\cite{igcn}.
\accname is able to take further benefit from those reordering schemes, due to two reasons.
First, when a graph is reordered, it forms small, dense clusters.
In consequence, the access pattern becomes similar to that of a smaller working set.
This would change the optimal number of vertex tiles, but \accname is able to adapt to it thanks to \atm. Second, reordering helps improve the locality of the accesses.
Because \accname is a technique that exploits the locality of accesses using caches, reordering often strengthens \accname.

\figurename~\ref{fig:igcn_perf} plots the performance benefit of \accname implemented with reordering of I-GCN~\cite{igcn} on three datasets. 
As depicted in the graphs, combining reordering and \accname consistently add extra speedup over using each scheme on its own.
On multi-engine setting, combination of the reordering and \accname adds 87.9\% improvement on top of I-GCN in geometric mean, and 4.22\% over original \accname in geometric mean.
On the small scale setting, the improvement is 55.4\% over I-GCN and 9.04\% over original \accname.

\input{eval/gnn_var}

\subsection{Performance on GCN Variants}
Although \accname focuses on acceleration of GCNs, there are many variants of GCNs~\cite{ginconv, graphsage} and \accname can be used to execute them.
We implement two popular GCN variants, GraphSAGE~\cite{graphsage} and GINConv~\cite{ginconv}, which pose slight modification to the aggregation phase. %\JL{MG: chk and fill what was different}
\figurename~\ref{fig:gnn_perf} plots the performance of GCNs.
Each result shows almost identical trend, because they share the same aggregation-combination structure.
On GraphSAGE, SnF gets 63.9\% speedup over VT in multi-engine setting and 40.7\% in small scale setting.
On the other hand, on GINConv, SnF gets 85.4\% in multi-engine and 64.3\% in small scale setting.
Because GraphSAGE applies sampling that minimize the number of edge access which reduces the chance for feature reuse, its performance gain are slightly lower than in GCN.
On the other hand, GINConv uses smaller set of edge weights (that is part of \A), and thus speedup from reading the features become more prominant, leading to a higher speedup than GCN.
  
\subsection{Sensitivity to Feature Width}
\label{sec:sensitivity}
\input{eval/single_F}
In this subsection, we provide sensitivity analyses of the \accname.
One insight from Eq.~\ref{eq:compare} is that speedup of feature slicing depends on $B_F$.
\figurename~\ref{fig:single_f} presents performance sensitivity to the feature widths in geometric mean. 
As indicated in the figure, the effectiveness of \accname is reduced with the layers of small feature widths which often appear in the later layers of GCNs.
However, the impact of such layers would be marginal due to short execution times.
Furthermore, the configuration found by \accname can be reused in the succeeding layers.
With the trend of wider feature width and deeper layers, the benefit of ATM is likely to strengthen over time.

\begin{comment}

\fig{fig:single_hit_ratio} (left) shows the cache miss rate, with respect to $B_F\times B_V$ observed for PK dataset.
For each $B_F \times B_V$, we tested all possible combinations and report the best combination.
The miss rate decreases as $B_F\times B_V$ becomes larger as the working set size decreases.
The miss rate drops for VT only and \accname are mostly similar, showing that our assumption for the cost model generally holds in that the miss rate is roughly a function of the working set size.

  \input{eval/single_sense}

\fig{fig:single_hit_ratio} (right) shows the memory access count breakdown that explains how the miss rates and repetition counts are translated into performance.
The green bars represent the feature accesses, and the purple bars represent the topology data accesses.
%With vertex tiling, the reduction in the memory access count according to the tiling degree increase is relatively small. 
%Even though the miss rate drops,
With vertex tiling, the repetition count being added to the output feature accesses offsets the gain from the miss rate drop and results in a diminishing return.
On contrary, with \accname, the miss rate drop is almost directly translated into memory access reduction as the repetition is mostly added to the smaller topology data. 

\end{comment}

\subsection{Sensitivity to System}
\input{eval/scale_new}
\figurename~\ref{fig:scale} shows how \accname performs with various number of engines and various memory systems, normalized at the performance of a small scale \accname with a DDR4-2666 channel.
With one DDR4-2666 channel peaked at 21.3GB/s, increasing the number of engines gives diminishing returns, as the execution becomes memory bound.
However, with memory subsystems with higher bandwidth, it requires more engines to fill the bandwidth.
For example, an HBM2 system that provides 256GB/s can gain from more processing engines, and exhibit almost linear speedup until eight engines.
Afterwards it reaches the saturation point.
Based on the study, we set the number of engines in the default setting to be eight.

\subsection{Sensitivity to Cache Size}
\input{eval/cache_size_sensi}

Analyzing the cache size sensitivity is essential because \accname takes advantage of efficient cache locality.
\figurename~\ref{fig:cachesensi} shows the performance of \accname for various cache sizes.
\accname generally outperforms other architectures.
\accname is especially advantageous in a small cache setting, so it works well with a smaller cache size to reduce silicon area.
In a small cache setting, the miss ratio of feature access, which \accname targets to minimize, increases, so \accname provides more speedup over other accelerators.
For example, in the 2MB setting, \accname shows 96.7\% improvement over VT, while the improvement of the 16MB setting is 73.1\% which is still a significant speedup.

\subsection{Sensitivity to Cache Configuration}
\input{eval/cachesense}
\figurename~\ref{fig:cachesense} shows the sensitivity to various cache configurations, including replacement policies, number of ways, and block size.
For replacement, we have tested random and RRIP~\cite{rrip}.  
Random replacement resulted in an surprisingly worse speedup, due to its inability to capture the locality. 
RRIP was able to gain a marginal amount of speedup, but it was not enough to justify the additional cost for storing more states.
While number of ways did not provide much changes, block sizes altered much performance.
When decreased, it becomes lower then the size of the DRAM burst, wasting a lot of the bandwidth.
When increased, the size exceeded the trivial spatial locality of the \fes, and the speedup degraded.

\subsection{Sensitivity on $B_F$ and $B_V$}
\label{sec:bf_sens}

\begin{figure}[t]
    \centering
    \includegraphics[width=\columnwidth]{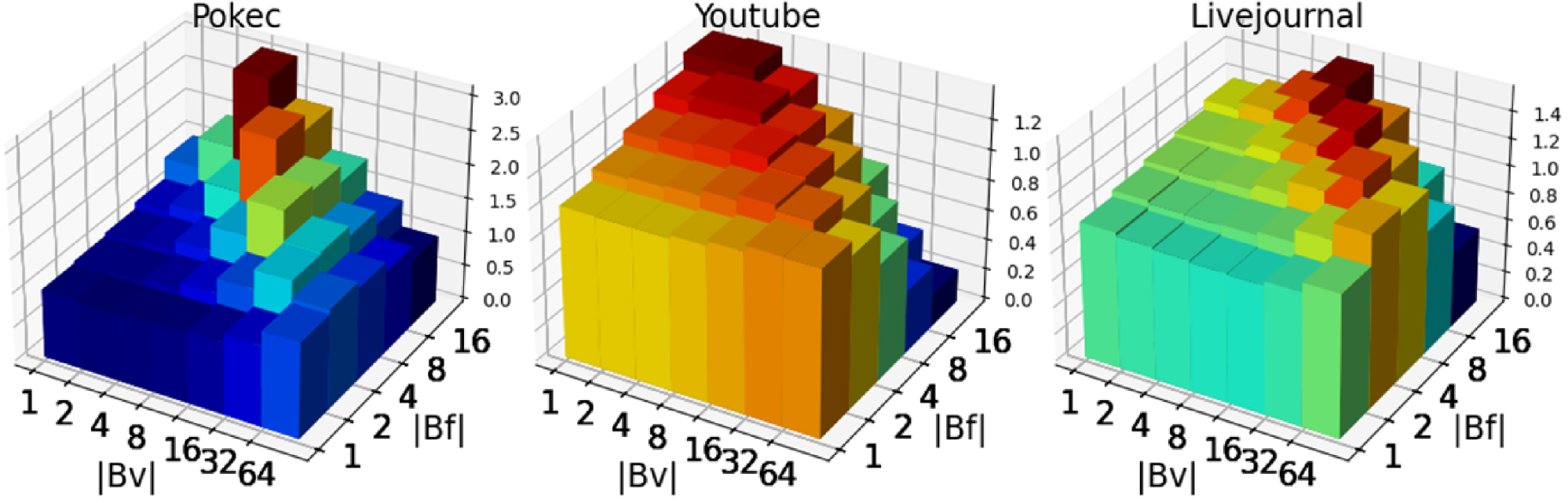}
    \caption{Sensitivity on $B_F$ and $B_V$.}
    \label{fig:bf_bv_sensi}
\end{figure}

\fig{fig:bf_bv_sensi} shows the sensitivity of \accname with respect to $B_F$ and $B_V$ on three selected datasets. 
The results exemplify the difficulty of tuning the number of vertex tiles $B_V$.
The optimal configuration is usually found on $B_F$=16 (the maximum), which supports our strategy to use feature slicing and then to adaptively find optimal $B_V$.
$B_V$, on the other side, is harder to tune since it depends too much on the individual graph topology.
Larger graphs favor larger $B_V$ because larger graphs would be harder to fit into caches.
% In our experiments, the optimal speedup was observed at $B_V$ is around 4.
% In ATM, we set the start $B_V$ to 4 for shortening the \textit{coarse\_morphing} phase.

 \input{eval/single_sense}
\fig{fig:single_hit_ratio} (left) shows the cache miss rate, with respect to $B_F\times B_V$ observed for PK dataset.
% For each $B_F \times B_V$, we tested all possible combinations and report the best combination.
The miss rate decreases as $B_F\times B_V$ becomes larger as the working set size decreases.
The miss rate drops for VT only and \accname are mostly similar, showing that our assumption for the cost model generally holds in that the miss rate is roughly a function of the working set size.

\fig{fig:single_hit_ratio} (right) shows the memory access count breakdown that explains how the miss rates and repetition counts are translated into performance.
The green bars represent the feature accesses, and the purple bars represent the topology data accesses.
%With vertex tiling, the reduction in the memory access count according to the tiling degree increase is relatively small. 
%Even though the miss rate drops,
In vertex tiling, the repetition count being added to the output feature accesses offsets the gain from the miss rate drop and results in a diminishing return.
However, with \accname, the miss rate drop is almost directly translated into memory access reduction as the repetition is added to smaller topology data. 

\subsection{Multi-Chip Module Scaling Results}

\input{eval/multi_chip_perf}

\begin{figure}
  \includegraphics[width=\columnwidth]{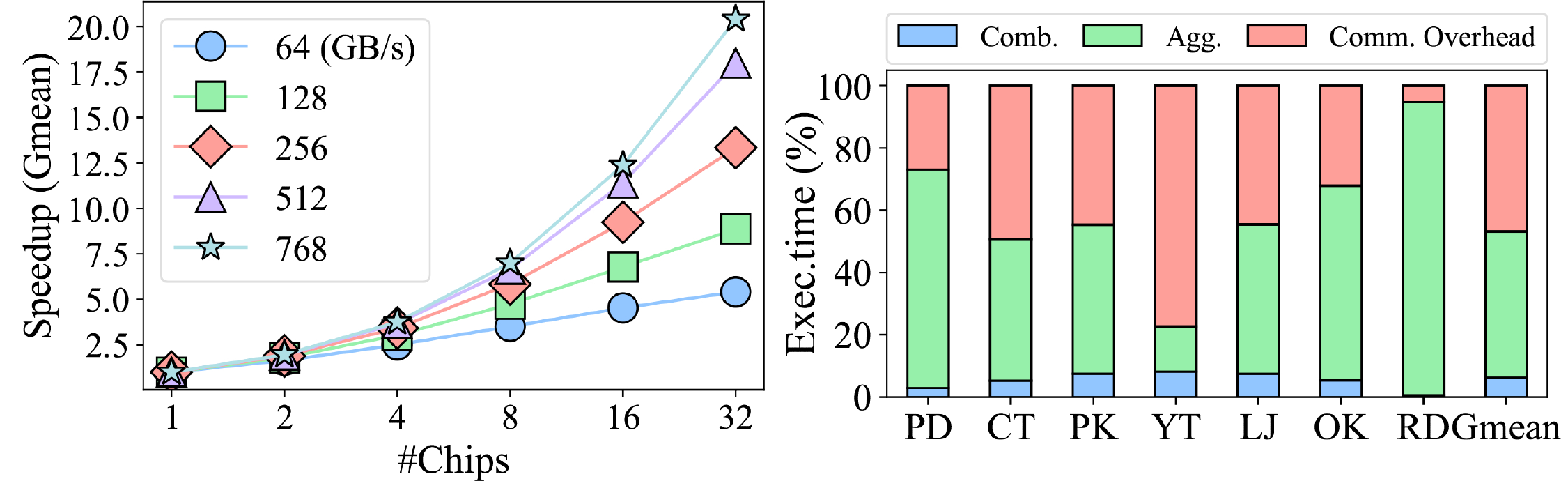}   
  \caption{Multi-chip scaling results (left) and execution time breakdown (right).}
  \vspace{-3mm}
\label{fig:multichip}
\end{figure}
\figurename~\ref{fig:multi_chip_perf} shows the multi-chip-module (MCM) accelerator performance where 16 chips in total form a module. 
Following other MCM accelerators~\cite{simba,mcmgpu} we use 256 GB/s bandwidth ring interconnect, with \SI{20}{\nano\second} per-hop latency. 
Even though all accelerators use the same strategy explained in Section~\ref{sec:mcm}, \accname achieves superior speedup over all others.
Further scaling results are provided in \figurename~\ref{fig:multichip}.
On the left, it plots the scalability according to the interconnect bandwidth.
Using interconnect with high bandwidth, it scales to some degree, but its speedup is limited for low-bandwidth interconnects.
On the right, the figure shows the breakdown of combination, aggregation and the communication overhead not hidden by the overlap.

The data with a large average degree such as PD, OK and RD scale well, because the communication time depends on the features per vertex and the computation time generally is proportional to the number of edges. 
On the other hand, low-degree graph such as YT shows poor scalability due to the low average degree and thus high communication overhead.

\subsection{Experiments on Small-Scale Datasets}

\input{eval/reproduce}

In addition to the datasets we evaluated, there exist some small-scale datasets often evaluated in some prior art~\cite{gcnax, awb, hygcn}: Cora, Citeseer~\cite{coraciteseer}, Pubmed~\cite{pubmed}, and Nell~\cite{nell}.
%~\cref{fig:reproduce} shows the experimental results on four additional small-scale datasets.
%These small-scale datasets are often too small, with only a few thousand vertices.
To make a fair comparison, we also report performance on those datasets in \figurename~\ref{fig:reproduce}.
We used 512 KB caches sizes for all architectures following \cite{gcnax}.
The performance improvement of \accname over the baseline is 106\%, which is much larger than the results \accname gets from large datasets, as depicted in \figurename~\ref{fig:reproduce_speed}.
The results align with the observations from previous work~\cite{awb, gcnax}.
However, we do not use these small-scale datasets for our main evaluation because these datasets often contain only a few thousands of vertices and thus consume tiny cycles to compute GCN compared to larger ones, as illustrated in \figurename~\ref{fig:reproduce_cycle}.
% ~\cref{fig:reproduce} is the experiments result on small-scale datasets that used our prior works. 
% Theses datasets are often too small (only a few thousand vertices).
% As depicted in~\cref{fig:reproduce_speed}, the performance improvement of \accname over VT is 2.06$\times$ and this result is much larger than the dataset that we target on.
% But, we do not deal with these on our main evaluation because they consume small cycles compared to large datasets. 
% ~\cref{fig:reproduce_cycle} is the actual cycle result on our simulator. the total compute cycle of small-scale datasets are too small compare to RD that has smallest number of vertices among our targeted dataset. So, we do not choice these for our target.

\section{Related Work}
\subsection{Accelerators for GCN and Graph Processing}
There are many existing architectural approaches that aim to accelerate GCNs.
%Earlier works try to accelerate GCNs, which is the baseline model of GNNs.
HyGCN~\cite{hygcn} proposed the hybrid computation architecture using the characteristic difference in the aggregation and the combination phase in GCNs.
GNNA~\cite{dacgcn} used a tiled architecture to accelerate GCN.
EnGN~\cite{engn} adopted a tiled topology data and gathered information about highly visited vertices to allocate some portion of cache for them to enhance cache efficiency.
Moreover, GCNAX~\cite{gcnax} uses a perfect tiling strategy by doing off-line analysis, which finds the best dataflow and tiling configuration. 
However, their perfect tiling over-estimates a required buffer size, leading to suboptimal performance.
In contrast, this paper focuses on cache efficiency and claims that \fes often provides a better efficiency than prior solutions.
AWB-GCN~\cite{awb} used column-based execution and gave insight into distributing workload.
Recent work also supports variants of GCNs that integrate the edge information or self-attention into GCNs.
\accname shows a similar speedup for such GCN variants because feature slicing and dynamic analysis technology can be easily applied without much challenge.

Not only GCNs, but classic graph processing has been an important application for decades, and there were various works which propose dedicated accelerators.
They utilized ASICs~\cite{graphicionado, ozdal, gramer}, FPGAs~\cite{extrav, polygraph, thousand}, or both~\cite{flexminer} to implement these accelerators.
Focusing on their memory-intensive nature, there were accelerators that utilize 3D-stacked memories~\cite{hmc} as a processing-in-memory accelerator~\cite{tesseract, graphq, graphp}, or modify flash storage to be friendly for graph processing.

Unfortunately, the computational pattern of such classic graph processing is distinct from that of the GCNs, because the features associated with each vertex is often shallow, with one or two elements per vertex. 
Because GCNs exhibit hybrid computational pattern and the feature vector associated with each vertex is relatively long, GCN accelerators require different design from that of the graph processing accelerators.

\subsection{Memory Aware Accelerators for DNNs}

DNN acceleration often accompanies memory-bound operations, so many works optimize memory bandwidth for successful speedup~\cite{Rhu2016vDNN, Hyun2020NeuMMU, Siu2018Memory}.
Some works precisely select the dataflow~\cite{shidiannao, eyeriss, tpu} or block data~\cite{overrated} to improve data reuse.
On-chip memory management is also crucial for optimization.
Many works try to efficiently deal with on-chip memory by scheduling, allocation, or sharing~\cite{overcoming, smartshuttle, splitcnn}.
DNNs often include of multiple layers, which can be utilized to optimize memory usage~\cite{fusedlayer, ecnn, mbs, bnff}.
This characteristic can be adopted not only for inference~\cite{fusedlayer, ecnn} and training~\cite{mbs} but also for batch normalization~\cite{bnff}.

These methods apply to the regular access pattern of conventional DNNs, which is not directly applicable to GCNs.
Therefore, GCN accelerator designs are different from conventional DNN accelerator designs because they target the irregular accesses.
%Thus, GCN accelerators differ from those in the techniques they adopt for dealing with the random accesses.

% Memories are known to be important factors of DNN accelerators, 
% and optimizations focused on memory bandwidth are often crucial to performance. 
% %One popular method is to optimize the on-chip memory usage.
% Many accelerators adopt different dataflows that determine which data should be reused~\cite{shidiannao, eyeriss, tpu}.
% %
% Some researchers have claimed that data blocking is a more important decision than the choice of dataflow~\cite{overrated}. 
% In a similar sense, some work has focused on the management of on-chip memories, such as allocation, sharing, or scheduling~\cite{overcoming, smartshuttle, splitcnn}.

% Another direction is to take advantage of the multi-layered nature of DNNs. 
% %By keeping inter-layer data as much as possible in on-chip memory, the bandwidth for off-chip memory can be reduced
% ~\cite{fusedlayer, ecnn, mbs, bnff}.
% These methods have been successfully applied to inference~\cite{fusedlayer, ecnn}, training~\cite{mbs}, and batch normalization~\cite{bnff}.

\begin{comment}
Among many differences, we highlight that these techniques usually handle on-chip memory as a manually managed scratchpad. 
This methods are valid because the memory access pattern in conventional DNNs is highly regular and fundamentally different from this work, which aims large volume of irregular accesses. 
\end{comment}

\subsection{Sparse DNN Accelerators}
Some DNN accelerators attempt to utilize sparsity in DNNs.
Computation pruning~\cite{Hua2019Boosting, Niu2020PatDNN, Silfa2019Neuron} was a popular target because it can easily apply to existing hardware designs.
Sparse operation support~\cite{Zhu2019Sparse, wang2021dual} or algebraic sparsity exploration~\cite{ringcnn} were also studied.
There are also zero-value aware designs , such as removing zero-value weights~\cite{han2016eie, Zhang2016CambriconX, Zhou2018CambriconS, Albericio2016Cnvlutin, Ding2017CirCNN} or considering zero-value activations~\cite{Parashar2017SCNN, Gondimalla2019SparTen, Judd2017Cnvlutin2, samsung}.
Some designs target SpMV~\cite{Sadi2019Efficient, Asgari2020ALRESCHA} and SpMM~\cite{sparch, hegde2019extensor, Qin2020SIGMA}, which can be used to exploit sparsity in DNNs.
In addition, some works~\cite{kanellopoulos2019smash, tensaurus, Pentecost2019MaxNVM, spzip} co-designed software and hardware for tensor compression.
However, the sparsity targeted from these work is often far below that of the GCNs, 
because the sparsity of graph topology is near 100\%, which is difficult for pruned DNNs or zero-valued activations to reach.
Therefore, despite the much work on sparse DNN accelerations, there exists a need for dedicated GCN accelerators.

\section{Conclusion}
We have proposed \saf, an accelerator design oriented at efficiently utilizing the caches for GCN accelerators.
With \fes, \saf greatly improves the effectiveness of the limited on-chip  cache compared to previous work.
Taking advantage of the fact that \fes yields multiple exact same computational patterns, we further propose \atm that dynamically configures the tiling for better cache efficiency.
Experimental results show that \saf achieves superior performance over the existing designs, and provides relaxation for the need for off-line analysis.

\begin{acks}
This work was partly supported by
% NRF
%  - 신진연구: 2022R1C1C1011307 (JH), 2022R1C1C1008131 (YS)
% IITP
%  - PIM SW: 2021-0-00853
%  - AI대학원: 2020-0-01361
the National Research Foundation of Korea (NRF) grants (2022R1C1C1011307, 2022R1C1C1008131) and
Institute of Information \& communications Technology Planning \& Evaluation (IITP) grants (2021-0-00853, 2020-0-01361) funded by the Korea government (MSIT).
% IDEC
The EDA tool was supported by the IC Design Education Center (IDEC), Korea.
% BK FOUR
Mingi Yoo, Jaeyong Song, Hyeyoon Lee, Jounghoo Lee, and Youngsok Kim are with the Department of Computer Science at Yonsei University and have been partly supported by the BK21 FOUR (Fostering Outstanding Universities for Research) funded by the Ministry of Education (MOE, Korea) and National Research Foundation of Korea (NRF).
\end{acks}
%%%%%%%%%%%%%%%%%%%%%%%%%%%%%%%%%%%%%%%%%%%%%%%%%%%%%%%%%%%%%%%%%%%%%%%%%%%%%%%%

\bibliographystyle{ACM-Reference-Format}
\bibliography{refs}
%\bibliography{sample-base}
%\bibliography{refs}

%\bibliographystyle{IEEEtranS}

\end{document}

%% file: packages.tex
\usepackage{amsmath,amsfonts}
\usepackage{nccmath} %medsize

\usepackage[utf8]{inputenc}
\usepackage{algorithm}
\usepackage[noend]{algpseudocode}
\usepackage{textcomp}

\usepackage[capitalize]{cleveref}
\usepackage{subcaption}
\usepackage{graphicx}
\usepackage{verbatim}
\usepackage{nicefrac}
\usepackage[binary-units=true]{siunitx}
\usepackage{float} % [H] option for figure

\usepackage[export]{adjustbox} % left, right for includegraphics

\usepackage{balance}

\usepackage{soul, color}

\usepackage{xcolor} 
\usepackage{booktabs}
\usepackage{multicol}
\usepackage{multirow}
\usepackage{makecell}
\usepackage{xspace}
\usepackage{marginnote}
\usepackage{setspace} % for setstretch

\usepackage{pgfplots}
\usepackage{pgfplotstable}
\usepgfplotslibrary{groupplots}
\usepackage{tikz}
\usetikzlibrary{patterns}
\usetikzlibrary{backgrounds}

\usepackage{comment}
\usepackage{marginnote}

\usepackage{listings}
\usepackage{pifont}% http://ctan.org/pkg/pifont
\newcommand{\cmark}{\color{olivegreen}\ding{51}}%
\newcommand{\xmark}{\color{red}\ding{55}}%

\usepackage[linewidth=1pt]{mdframed}
\usepackage{tcolorbox}

\usepackage{enumitem}

%% file: commands.tex
\definecolor{olivegreen}{rgb}{0, 0.6, 0}
\definecolor{grannysmithapple}{rgb}{0.66, 0.89, 0.63}
\definecolor{ceruleanblue}{rgb}{0.16, 0.32, 0.75}
\definecolor{babyblue}{rgb}{0.54, 0.81, 0.94}
	
\newcommand{\JL}[1]{{\color{red}[\textbf{\sc JLee}: \textit{#1}]}}
\definecolor{blue(ncs)}{rgb}{0.0, 0.53, 0.74}
\definecolor{blush}{rgb}{0.87, 0.36, 0.51}
\newcommand{\NK}[1]{{\color{blue(ncs)}[\textbf{\sc NK}: \textit{#1}]}}

\definecolor{carminered}{rgb}{1.0, 0.0, 0.22}
\newcommand{\MG}[1]{{\color{carminered}[\textbf{\sc Mingi}: \textit{#1}]}}

\newcommand{\JS}[1]{{\color{blue(ncs)}[\textbf{\sc JS}: \textit{#1}]}}

\renewcommand{\JS}[1]{}
\newcommand{\rev}[1]{{\color{olivegreen}#1}}
\newcommand{\revn}[1]{{\color{olivegreen}#1}}

\def\final{}   % uncomment this for submission version
\ifdefined\final
\renewcommand{\JL}[1]{}
\renewcommand{\NK}[1]{}
\renewcommand{\MG}[1]{}
\renewcommand{\rev}[1]{#1}
\fi

\newcommand{\SaF}{Slice-and-Forge\xspace}
\newcommand{\saf}{slice-and-forge\xspace}

\newcommand{\atm}{automatic tile morphing\xspace}
\newcommand{\Atm}{Automatic tile morphing\xspace}
\newcommand{\ATM}{Automatic Tile Morphing\xspace}

\newcommand{\fes}{{feature slicing}\xspace}
\newcommand{\Fes}{{Feature slicing}\xspace}
\newcommand{\FeS}{{Feature Slicing}\xspace}

\newcommand{\accname}{SnF\xspace} %Slice and Forge 
\newcommand{\accnameplus}{SnF} %Slice and Forge 

\newcommand{\fig}[1]{\figurename~\ref{#1}}

\newcommand*\circled[1]{\tikz[baseline=(char.base)]{
            \node[shape=circle,draw,inner sep=0.4pt] (char) {#1};}}

\newcommand{\A}{$\tilde{A}$\xspace}

\newcommand{\revref}[1]{\hyperref[rev:#1]{\color{magenta}#1}}
\newcommand{\figref}[1]{\hyperref[fig:#1]{\color{blue}\figurename~\ref{fig:#1}}}
\newcommand{\tblref}[1]{\hyperref[tbl:#1]{\color{blue}\tablename~\ref{tbl:#1}}}
\newcommand{\algoref}[1]{\hyperref[alg:#1]{\color{blue}Algorithm~\ref{alg:#1}}}

\usepackage{soul}

\definecolor{black}{HTML}{000000}
\definecolor{white}{HTML}{ffffff}
\definecolor{color1}{HTML}{90ee90}
\definecolor{color2}{HTML}{F0E68C}
\definecolor{color3}{HTML}{DCD0FF}
\definecolor{color4}{HTML}{B19CD9}
\definecolor{color5}{HTML}{FFB6C1}
\definecolor{color6}{HTML}{20B2AA}
\definecolor{color7}{HTML}{87CEEB}
\definecolor{color8}{HTML}{FFA07A}

\pgfplotsset{compat=newest}
\makeatletter
\newcommand\resetstackedplots[1]{
\makeatletter
\pgfplots@stacked@isfirstplottrue
\makeatother
\pgfplotstablenew[
  create on use/x/.style={create col/expr={\pgfplotstablerow}},
  create on use/y/.style={create col/expr={0}},
  columns={x,y}]{#1}\zerotable
\addplot [forget plot,draw=none] table[x=x,y=y] from \zerotable;
}
\makeatother

%% file: eval/moti_vsense.tex
\begin{figure}[t]
    \begin{tikzpicture}%[framed,background rectangle/.style={ultra thick,draw=olivegreen}]
    
\pgfplotstableread{

id	BF=2	pk2	pk8	pd2	pd8	ct2	ct8	pk2best	pk8best	ct2best	ct8best
1	VT1	1.025887978	1.206295733	1.047614882	1	1.31716466	1.103208986	nan	nan	1.31716466	nan
2	VT2	1.002412258	1.270405296	1.037521176	1.138355741	1.242652038	1.090660528	nan	nan	nan	nan
3	VT4	1	1.485106546	1.024877148	1.306784475	1.154474348	1.09378678	nan	nan	nan	nan
4	VT8	1.017589164	1.880131837	1.005459085	1.658056438	1.074716673	1.115465296	nan	1.880131837	nan	nan
5	VT16	1.094326732	1.828331593	1	2.052623216	1.031970932	1.175317421	nan	nan	nan	nan
6	VT32	1.312757514	1.419262473	1.042034871	1.698950185	1	1.281294642	nan	nan	nan	1.281294642
7	VT64	1.390160769	1	1.129508969	1.362472005	1.111157344	1	1.390160769	nan	nan	nan
}\data

\pgfplotscreateplotcyclelist{custom}{%https://tex.stackexchange.com/questions/134346/different-marker-shape-for-pgf-tikz
solid, thick, mark size=2pt, every mark/.append style={solid, thick, fill=color3}, mark=diamond*\\%
dotted, thick, mark size=1.5pt, every mark/.append style={solid, thick, fill=color7}, mark=square*\\%
solid, thick, mark size=3pt, every mark/.append style={solid, thick, color=red}, mark=o\\%
dotted, thick, mark size=3pt, every mark/.append style={solid, thick, color=red}, mark=o\\%
densely dotted, thick, mark size=3pt, every mark/.append style={solid, fill=color3}, mark=diamond*\\%
loosely dotted, thick, mark size=3pt, every mark/.append style={solid, fill=color5}, mark=triangle*\\%
solid, thick, mark size=3pt, every mark/.append style={solid, thick, fill=gray}, mark=*\\%
dashed, thick, every mark/.append style={solid, fill=color5},mark=otimes*\\%
loosely dashed, thick, every mark/.append style={solid, fill=gray},mark=*\\%
densely dashed, thick, every mark/.append style={solid, fill=gray},mark=square*\\%
dashdotted, thick, every mark/.append style={solid, fill=gray},mark=otimes*\\%
dashdotdotted, thick, every mark/.append style={solid},mark=star\\%
densely dashdotted,thick, every mark/.append style={solid, fill=gray},mark=diamond*\\%
}

\begin{groupplot}[group style={vertical sep=2.8em,horizontal sep=3em,group size= 2 by 1},height=3.6cm,width=0.55\columnwidth]

\nextgroupplot[
%font=\footnotesize\rmfamily,
%title style={yshift=-1.5mm},
	height=3cm,
	ymin=0,ymax=2,
	xticklabels={1, 2, 4, 8, 16, 32, 64, 128},
	x label style = {font=\scriptsize},
	xtick=data, x tick label style={font=\footnotesize,
	yshift=0.5mm}, tick style = transparent,
	xlabel= Number of Vertex Tiles (\textbf{Pokec}),
    ytick={0,0.5,1,1.5,2},
	y tick label style={font=\footnotesize},
	ymajorgrids=true, major grid style={thin,dashed},
    ylabel= Norm. Speedup,
    ylabel style = {font=\scriptsize, yshift=-1mm},
	%axis x line*=bottom,
	%axis y line*=none,
	%legend style={draw=none, fill=none, at={(0.69,0.5), font=\tiny},
	cycle list name=custom,
	legend cell align={left},
	legend cell align={left},
	legend style={draw=none,  fill=none, at={(0.43,1.0), font=\scriptsize},
	anchor=south,legend columns=3,
	/tikz/every even column/.append style={column sep=0.2cm}
    }
]

\addplot table  [x=id, y=pk8] {\data};\addlegendentry{$|F|=32$}
\addplot table  [x=id, y=pk2] {\data};\addlegendentry{$|F|=128$}
\addplot +[unbounded coords=discard] table  [x=id, y=pk8best] {\data};
\addplot +[unbounded coords=discard] table  [x=id, y=pk2best] {\data};

\nextgroupplot[
%font=\footnotesize\rmfamily,
%title style={yshift=-1.5mm},
	 height=3cm,
	ymin=0,ymax=1.7,
	xticklabels={1, 2, 4, 8, 16, 32, 64, 128},
	x label style = {font=\scriptsize},
	xtick=data, x tick label style={font=\footnotesize,
	yshift=0.5mm}, tick style = transparent,
	xlabel= Number of Vertex Tiles (\textbf{Citation}),
    ytick={0,0.5,1,1.5,2},
	y tick label style={font=\footnotesize},
	ymajorgrids=true, major grid style={thin,dashed},
    ylabel= Norm. Speedup,
    ylabel style = {font=\scriptsize, yshift=-1mm},
	%axis x line*=bottom,
	%axis y line*=none,
	%legend style={draw=none, fill=none, at={(0.69,0.5), font=\tiny},
	cycle list name=custom,
	legend cell align={left},
	legend style={draw=none, fill=none, at={(0.43,1.0), font=\scriptsize},
	anchor=south,legend columns=3,
	/tikz/every even column/.append style={column sep=0.2cm}}
]

\addplot table  [x=id, y=ct8] {\data};
\addplot table  [x=id, y=ct2] {\data};
\addplot +[unbounded coords=discard] table  [x=id, y=ct8best] {\data};
\addplot +[unbounded coords=discard] table  [x=id, y=ct2best] {\data};

\end{groupplot}
\end{tikzpicture}
 \caption{Sensitivity of GCN to number of vertex tiles. The best performing number of vertex tiles are marked with {\textcolor{red}{red}} circles.}
 %\vspace{-5mm}
 \label{fig:moti}
%  \vspace{-3mm}
  \end{figure}

%% file: eval/single_perf_side.tex
\begin{figure}

%\vspace{-2mm}

\begin{subfigure}[t]{\columnwidth}
\input{eval/multi_perf}
    \caption{Multi-engine setting.}
    \label{fig:multi_perf}
\end{subfigure}

\begin{subfigure}[t]{\columnwidth}
    \begin{tikzpicture}

%fake engn values for pk and lj and geomean	

\centering
%\begin{groupplot}[group style={vertical sep=2.8em,horizontal sep=3em,group size= 2 by 1},height=3.6cm,width=1.08\columnwidth]

\begin{axis} %nextgroupplot
[
	ybar=0pt, %this value determines the space between bars
	width=1.08\columnwidth, 
	height=3cm,
	bar width=2.5pt, 
	ymin=0,ymax=2.6,
	xticklabels={LJ, RD, PK, YT, PD, CT, OK, Gmean},
	xtick=data, x tick label style={font=\footnotesize, yshift=1.5mm}, tick style = transparent,
	ytick={0,1.0,...,4.0},
	y tick label style={font=\footnotesize},
	ymajorgrids=true, major grid style={thin,dashed},
    ylabel={Speedup},
    ylabel style = {font=\scriptsize, yshift=-1mm},
	%axis x line*=bottom,
	%axis y line*=none,
	%legend style={draw=none, fill=none, at={(0.69,0.5), font=\tiny},
	%legend style={draw=none, fill=none, at={(1.08,1.05), font=\scriptsize},
	legend cell align={left},
	legend style={draw=none, fill=none, at={(.47,1.05), font=\scriptsize},
	anchor=south,legend columns=4,
	/tikz/every even column/.append style={column sep=.2cm}},
    legend image code/.code={%
      \draw[#1] (0cm,-0.1cm) rectangle (0.2cm,0.1cm);
    }
	]

\addplot[fill=color2] table [x=id, y=VT] {eval/single_16MB.txt};  %\addlegendentry{VT}
\addplot[fill=color1] table [x=id, y=Base] {eval/single_16MB.txt};  %\addlegendentry{HyGCN}
\addplot[fill=color7] table [x=id, y=ENGN] {eval/single_16MB.txt};  %\addlegendentry{EnGN}
\addplot[fill=color6] table [x=id, y=AWB] {eval/single_16MB.txt};  %\addlegendentry{AWB-GCN}
\addplot[fill=color8] table [x=id, y=GCNAX] {eval/single_16MB.txt};  %\addlegendentry{GCNAX}
%\addplot[fill=color3] table [x=id, y=FS] {eval/single_16MB.txt};  \addlegendentry{\accname{}\textsubscript{Optimal}}
%\addplot[fill=color3] table [x=id, y=FS+VT] {eval/single_16MB.txt};  %\addlegendentry{\accname{}\textsubscript{Optimal}}
\addplot[fill=color4] table [x=id, y=ATM] {eval/single_16MB.txt};  %\addlegendentry{\accname{}}
% \addplot[fill=color6] table [x=id, y=ATM(improve)] {eval/single_16MB.txt};
% \addlegendentry{VT+\accname{}\textsubscript{Dynamic(improve)}}

\end{axis}	

\end{tikzpicture}
    \caption{Small-scale setting.}
    \label{fig:single_perf}
\end{subfigure}

\caption{Performance of \accname in (a) multi-engine setting and (b) small-scale setting.}
 \label{fig:perf}
  \end{figure}

%% file: eval/multi_perf.tex
%\begin{figure}[tp]

%\vspace{-2mm}

    \begin{tikzpicture}

%fake engn values for pk and lj and geomean			
\pgfplotstableread{
id	data	VT	Base	EnGN	AWB	GCNAX	LAC
4	Livejournal	1	1	1.009472007	1.448531808	1.291045225	1.754852074
6	Reddit	1	0.4974184485	1.000228268	1.017269283	1.153795946	1.211258802
2	Pokec	1	1	1.015564284	1.529729582	1.272892802	2.026013192
3	Youtube	1	1	1.048330988	1.467093934	0.9211969138	1.594156448
0	Products	1	1	1.034753339	1.755902927	1.2439966	2.76030455
1	Citation2	1	1	1.002697732	1.061807477	1.009463366	1.096764947
5	Orkut	1	0.9976735077	1.071864549	1.375787483	1.245830818	2.24197206
7	Geomean	1	0.9047530315	1.025836159	1.357212158	1.154433936	1.731154855

}\datatable

\begin{axis}[
	ybar=0pt, %this value determines the space between bars
	width=1.07\columnwidth, height=3cm,
	bar width=2.5pt, 
	ymin=0,ymax=3,
	xticklabels={LJ, RD, PK, YT, PD, CT, OK, Gmean},
	xtick=data, x tick label style={font=\footnotesize, yshift=1.5mm}, tick style = transparent,
	ytick={0,1,2},
	y tick label style={font=\footnotesize},
	ymajorgrids=true, major grid style={thin,dashed},
    ylabel={Speedup},
    ylabel style = {font=\scriptsize, yshift=-1mm},
	%axis x line*=bottom,
	%axis y line*=none,
	%legend style={draw=none, fill=none, at={(0.69,0.5), font=\tiny},
	legend cell align={left},
	legend style={draw=none, fill=none, at={(0.48,1.0), font=\scriptsize},
	anchor=south,legend columns=3,
	/tikz/every even column/.append style={column sep=0.5cm}},
    legend image code/.code={%
      \draw[#1] (0cm,-0.1cm) rectangle (0.2cm,0.1cm);
    }
	]  

\addplot[fill=color2] table [x=id, y=VT] {\datatable};       \addlegendentry{VT}
\addplot[fill=color1] table [x=id, y=Base] {\datatable};  \addlegendentry{HyGCN}
\addplot[fill=color7] table [x=id, y=EnGN] {\datatable};  \addlegendentry{EnGN}
\addplot[fill=color6] table [x=id, y=AWB] {\datatable};  \addlegendentry{AWB-GCN}
\addplot[fill=color8] table [x=id, y=GCNAX] {\datatable};\addlegendentry{GCNAX}
%\addplot[fill=color3] table [x=id, y=SAF] {\datatable};  \addlegendentry{\accname}
\addplot[fill=color4] table [x=id, y=LAC] {\datatable};  \addlegendentry{\accname (Proposed)}

\end{axis}
\end{tikzpicture}

%\caption{\rev{Performan/e of the multi-engine accelerator.}}
%\label{fig:multi_chip_perf}
% \vspace{-4mm}
%\end{figure}

%% file: eval/single_E.tex
\begin{figure}

% \vspace{-2mm}

    \begin{tikzpicture}
\pgfplotstableread{

id	data	AggB	AggF	CacheB	CacheF	DRAMB	DRAMF	dummysolid	dummypattern
0	PD	0.3316686161	0.3765915701	0.42841448	0.1548662278	0.2399169039	0.1281231269	0	0
1	CT	0.318249542	0.3706216524	0.4332025198	0.3065895614	0.2485479382	0.2432037478	0	0
2	PK	0.3431148227	0.3917343847	0.4184430211	0.2049749951	0.2384421563	0.1654045744	0	0
3	YT	0.3974990306	0.4605359676	0.3544076459	0.2966757506	0.2480933235	0.2254872455	0	0
4	LJ	0.3835475353	0.4459683332	0.3898910233	0.3597808526	0.2265614414	0.2262366469	0	0
5	OK	0.3582045593	0.3988852069	0.4089570711	0.1156833725	0.2328383695	0.120466449	0	0
6	RD	0.3753637985	0.4230822556	0.4007852345	0.02327960638	0.223850967	0.03228267524	0	0

}\datatable
%fake engn values for pk and lj and geomean			

\begin{axis}[
	ybar stacked, 
	width=\linewidth, height=3cm,
	bar width=8pt, 
	ymin=0,ymax=1.15,
	xticklabels={PD, CT, PK, YT, LJ, OK, RD},
	xtick=data, x tick label style={font=\footnotesize, yshift=0.5mm}, tick style = transparent,
    ytick={0,0.5,...,1.5},
	y tick label style={font=\footnotesize},
	ymajorgrids=true, major grid style={thin,dashed},
    % ylabel= Energy (\SI{}{\milli\joule}),
    ylabel= Normalized Energy,
    ylabel style = {font=\scriptsize, yshift=-1mm},
	%axis x line*=bottom,
	%axis y line*=none,
	%legend style={draw=none, fill=none, at={(0.69,0.5), font=\tiny},
	legend cell align={left},
	legend style={draw=none, fill=none, at={(0.48,1.0), font=\scriptsize},
	anchor=south,legend columns=3,
	/tikz/every even column/.append style={column sep=0.5cm}},
    legend image code/.code={%
      \draw[#1] (0cm,-0.1cm) rectangle (0.2cm,0.1cm);
    }
	]  

\addplot[fill=color5, xshift=4.4pt] table [x=id, y=AggF ] \datatable; \addlegendentry{Computation}
\addplot[fill=color1, xshift=4.4pt] table [x=id, y=CacheF ] \datatable; \addlegendentry{Cache}
\addplot[fill=color2, xshift=4.4pt] table [x=id, y=DRAMF ] \datatable; \addlegendentry{DRAM}

%dummies for legend
\addplot[fill=lightgray, 
        postaction={pattern color=black, pattern=north east lines}, xshift=4.4pt] table [x=id, y=dummypattern] \datatable; \addlegendentry{Baseline}
\addplot[fill=lightgray, xshift=4.4pt] table [x=id, y=dummysolid] \datatable; \addlegendentry{\accname}

\resetstackedplots{7}
\addplot[fill=color5,
        postaction={pattern color=black, pattern=north east lines}, 
        xshift=-3.5pt] table [x=id, y=AggB ] \datatable; %\addlegendentry{Comb}
\addplot[fill=color1,
        postaction={pattern color=black, pattern=north east lines}, 
        xshift=-3.5pt] table [x=id, y=CacheB ] \datatable; %\addlegendentry{Comb}
\addplot[fill=color2,
        postaction={pattern color=black, pattern=north east lines}, 
        xshift=-3.5pt] table [x=id, y=DRAMB ] \datatable; %\addlegendentry{Agg}

%\addplot[fill=color1] table [x=id, y=Comb] {\datatable};\addlegendentry{Baseline}
%\addplot[fill=color2] table [x=id, y=Agg] {\datatable};  \addlegendentry{AWB-GCN}
%\addplot[fill=color3] table [x=id, y=Cache] {\datatable};  \addlegendentry{EnGN}
%\addplot[fill=color4] table [x=id, y=Mem] {\datatable};  \addlegendentry{GCNAX} %fake

\end{axis}
\end{tikzpicture}

 \caption{Normalized energy consumption breakdown of baseline (shaded) and \accname (solid).}
 \label{fig:single_E}
  \end{figure}

%% file: eval/igcn_perf.tex
\begin{figure}

    \begin{tikzpicture}%[framed,background rectangle/.style={ultra thick,draw=olivegreen}]
    \pgfplotsset{
every axis title/.append style={at={(0.5,-0.7)},font=\footnotesize}
}

\pgfplotstableread{
id	data	HyGCN HyGCN+IGCN SnF SnF+IGCN
0	Citation	1	1.539395009	1.77278434	1.919054578
1	Pokec	1	1.092799877	1.376985388	1.43539818
2	Youtube	1	1.0091466	1.360730099	1.36494326
3	Geomean	1	1.192930084	1.492058129	1.554979872
}\igcntableA
%fake values

\pgfplotstableread{

id	data2	HyGCN HyGCN+IGCN SnF SnF+IGCN
0	Citation	1	1.156755927	1.24883996	1.505344838
1	Pokec	1	1.41127911	2.072401338	2.206216239
2	Youtube	1	1.651086867	1.975637629	1.996067545
3	Geomean	1	1.391686848	1.722777985	1.878537069
}\igcntableB

\begin{groupplot}[group style={vertical sep=2.8em,horizontal sep=1em,group size= 2 by 1},height=3.6cm,width=0.58\columnwidth]

\nextgroupplot[title=Multi-engine,
	ybar=0pt, %this value determines the space between bars
	height=3cm,
	bar width=3.5pt, 
	ymin=0,ymax=2.5,
	xticklabels={CT, PK, YT, Gmean},
	xtick=data, x tick label style={font=\footnotesize, yshift=1.5mm}, tick style = transparent,
	ytick={0,0.5,1,1.5,2,2.5},
    %yticklabel={\pgfmathparse{\tick*100}\pgfmathprintnumber{\pgfmathresult}\%},
	%y tick label style={font=\footnotesize},
		ymajorgrids=true, major grid style={thin,dashed},
    ylabel={Speedup},
    ylabel style = {font=\scriptsize, yshift=-2mm},
	%axis x line*=bottom,
	%axis y line*=none,
	%legend style={draw=none, fill=none, at={(0.69,0.5), font=\tiny},
	legend style={draw=none, fill=none, at={(1.1,1.0), font=\scriptsize},
	anchor=south,legend columns=8,
	/tikz/every even column/.append style={column sep=.5cm}},
    legend image code/.code={%
      \draw[#1] (0cm,-0.1cm) rectangle (0.2cm,0.1cm);
    }
	]  

\addplot[fill=color1] table [x=id, y=HyGCN] {\igcntableB};\addlegendentry{HyGCN}
\addplot[fill=color2] table [x=id, y=HyGCN+IGCN] {\igcntableB};  \addlegendentry{I-GCN}
\addplot[fill=color3] table [x=id, y=SnF] {\igcntableB};  \addlegendentry{SnF}
\addplot[fill=color4] table [x=id, y=SnF+IGCN] {\igcntableB};  \addlegendentry{SnF+I-GCN}

\nextgroupplot[title=Small scale,
	ybar=0pt, %this value determines the space between bars
	height=3cm,
	bar width=3.5pt, 
	ymin=0,ymax=2.5,
	xticklabels={CT, PK, YT, Gmean},
	xtick=data, x tick label style={font=\footnotesize, yshift=1.5mm}, tick style = transparent,
	ytick={0,0.5,1,1.5,2,2.5},
	y tick label style={font=\footnotesize},
	yticklabel={\empty},
	ymajorgrids=true, major grid style={thin,dashed},
    ylabel style = {font=\scriptsize, yshift=-.5mm},
	%axis x line*=bottom,
	%axis y line*=none,
	%legend style={draw=none, fill=none, at={(0.69,0.5), font=\tiny},
	legend style={draw=none, fill=none, at={(1.1,1.0), font=\scriptsize},
	anchor=south,legend columns=8,
	/tikz/every even column/.append style={column sep=.2cm}}, %even: test, odd: square box
    legend image code/.code={%
      \draw[#1] (0cm,-0.1cm) rectangle (0.2cm,0.1cm);
    }
	]  

\addplot[fill=color1] table [x=id, y=HyGCN] {\igcntableA};%\addlegendentry{HyGCN}
\addplot[fill=color2] table [x=id, y=HyGCN+IGCN] {\igcntableA};  %\addlegendentry{I-GCN}
\addplot[fill=color3] table [x=id, y=SnF] {\igcntableA};  %\addlegendentry{SnF}
\addplot[fill=color4] table [x=id, y=SnF+IGCN] {\igcntableA};  %\addlegendentry{SnF+I-GCN}

\end{groupplot}
\end{tikzpicture}

 \caption{Performance of I-GCN reordering.}
 \vspace{-3mm}
 \label{fig:igcn_perf}
  \end{figure}

%% file: eval/gnn_var.tex
\begin{figure}

    \begin{tikzpicture}%[framed,background rectangle/.style={ultra thick,draw=olivegreen}]
        \pgfplotsset{
every axis title/.append style={at={(0.5,-0.7)},font=\footnotesize}
}
\pgfplotstableread{
id	data	Base	VT	FS	FS+VT	ATM	AWB	ENGN GCNAX
0	GCN	1	0.9616457591	1.12611746	1.54746055	1.490720289	1.222574064	1.083479883	1.391152026
1	GraphSAGE	1	0.9949370048	1.284879295	1.421362689	1.40790132	1.223931301	1.053362172	1.305474518
2	GINConv	1	0.957486871	1.222566107	1.672687439	1.642634313	1.319177243	1.03482279	1.527802645
}\gnnstableA

\pgfplotstableread{

 id	data	VT HyGCN GCNAX EnGN AWB SAF LAC
0	GCN	1	0.9996118749	1.167926173	1.023293815	1.34698664	1.092432686	1.716529442
1	GraphSage	1	0.9991544003	1.225506795	1.029212379	1.369404571	1.043695912	1.638576228
2	GINConv	1	0.9996112249	1.157214016	1.030894459	1.446131358	1.1198742	1.853965453
}\gnnstableB

\begin{groupplot}[group style={vertical sep=2.8em,horizontal sep=1em,group size= 2 by 1},height=3.6cm,width=0.58\columnwidth]

\nextgroupplot[title=Multi-engine,
	ybar=0pt, %this value determines the space between bars
	enlarge x limits=.25,
	height=3cm,
	bar width=3.5pt, 
	ymin=0,ymax=2,
	xticklabels={GCN, GraphSAGE, GINConv},
	xtick=data, x tick label style={font=\scriptsize, yshift=1.5mm}, tick style = transparent,
	ytick={0,0.5,1,1.5,2},
	y tick label style={font=\footnotesize},
	ymajorgrids=true, major grid style={thin,dashed},
    ylabel={Speedup},
    ylabel style = {font=\scriptsize, yshift=-.5mm},
	%axis x line*=bottom,
	%axis y line*=none,
	%legend style={draw=none, fill=none, at={(0.69,0.5), font=\tiny},
	legend style={draw=none, fill=none, at={(1.1,1.0), font=\scriptsize},
	anchor=south,legend columns=8,
	/tikz/every even column/.append style={column sep=.1cm}}, %even: test, odd: square box
    legend image code/.code={%
      \draw[#1] (0cm,-0.1cm) rectangle (0.2cm,0.1cm);
    }
	]

\addplot[fill=color2] table [x=id, y=VT] {\gnnstableB};       \addlegendentry{VT}
\addplot[fill=color1] table [x=id, y=HyGCN] {\gnnstableB};  \addlegendentry{HyGCN}
\addplot[fill=color7] table [x=id, y=EnGN] {\gnnstableB};  \addlegendentry{EnGN}
\addplot[fill=color6] table [x=id, y=AWB] {\gnnstableB};  \addlegendentry{AWB-GCN}
\addplot[fill=color8] table [x=id, y=GCNAX] {\gnnstableB};\addlegendentry{GCNAX}
% \addplot[fill=color3] table [x=id, y=SAF] {\datatable};  \addlegendentry{\accname}
\addplot[fill=color4] table [x=id, y=LAC] {\gnnstableB};  \addlegendentry{\accnameplus}

\nextgroupplot[title=Small scale,
	ybar=0pt, %this value determines the space between bars
	enlarge x limits=.25,
	 height=3cm,
	bar width=3.5pt, 
	ymin=0,ymax=2,
	xticklabels={GCN, GraphSAGE, GINConv},
	xtick=data, x tick label style={font=\scriptsize, yshift=1.5mm}, tick style = transparent,
	ytick={0,0.5,1,1.5,2},
	y tick label style={font=\footnotesize},
	ymajorgrids=true, major grid style={thin,dashed},
    yticklabel={\empty},
    ylabel style = {font=\scriptsize, yshift=-2mm},
	%axis x line*=bottom,
	%axis y line*=none,
	%legend style={draw=none, fill=none, at={(0.69,0.5), font=\tiny},
	legend style={draw=none, fill=none, at={(1.1,1.0), font=\scriptsize},
	anchor=south,legend columns=8,
	/tikz/every even column/.append style={column sep=.5cm}},
    legend image code/.code={%
      \draw[#1] (0cm,-0.1cm) rectangle (0.2cm,0.1cm);
    }
	]  
\addplot[fill=color2] table [x=id, y=VT] {\gnnstableA};     %\addlegendentry{VT}
\addplot[fill=color1] table [x=id, y=Base] {\gnnstableA};%\addlegendentry{HyGCN}
\addplot[fill=color7] table [x=id, y=ENGN] {\gnnstableA};%\addlegendentry{EnGN}
\addplot[fill=color6] table [x=id, y=AWB] {\gnnstableA};%\addlegendentry{AWB-GCN}
\addplot[fill=color8] table [x=id, y=GCNAX] {\gnnstableA};%\addlegendentry{GCNAX}
%\addplot[fill=color3] table [x=id, y=FS] {\datatable};  \addlegendentry{\accname{}\textsubscript{Optimal}}
% \addplot[fill=color4] table [x=id, y=FS+VT] {\datatable};  \addlegendentry{\accname{}\textsubscript{Optimal}}
\addplot[fill=color4] table [x=id, y=ATM] {\gnnstableA};  %\addlegendentry{\accname{}}

\end{groupplot}
\end{tikzpicture}

 \caption{Performance of variant GNN application.}
 \label{fig:gnn_perf}
  \end{figure}

%% file: eval/single_F.tex
   
\begin{figure}[t]
    \begin{tikzpicture}
\pgfplotstableread{

id	data	Base	VT	FS	FS+VT	ATM	AWB	ENGN GCNAX
% 0	32	1.00	1.23	1.02	1.33	1.07	1.16	1.30 1 
% 1	64	1.00	1.19	1.04	1.28	1.13	1.20	1.25 1
% 2	128	1.00	1.12	1.09	1.32	1.29	1.23	1.19 1
% 3	192	1.00	1.07	1.08	1.34	1.28	1.26	1.15 1
% 4	256	1.00	1.09	1.26	1.51	1.46	1.26	1.16 1
% 0	32	0.8925881073	1	0.8936863713	1.070427272	0.8985825826	0.9961606595	1.000096713	1.024698403
0	64	0.8565873518	1	0.9467065854	1.146669023	1.002174747	1.035199091	1.000095758	1.053783326
1	128	0.8102416364	1	0.9664462094	1.263912427	1.100758422	1.026474226	1.000093947	1.091285076
2	192	0.8037298352	1	1.112910049	1.422769463	1.3551144	1.136977433	1.000072275	1.251801372
3	256	0.9616457591	1   1.12611746	1.54746055	1.490720289	1.222574064	1.083479883	1.391152026
4	512 0.9988788685	1	1.312827627	1.738852625	1.698395292	1.404951625	1.0788279	1.441194721
}\datatable

\begin{axis}[
	ybar=0pt,
	enlarge x limits=0.15,
	width=\linewidth, height=3cm,
	bar width=3.5pt, 
	ymin=0,ymax=2.0,
	xticklabels={64, 128, 192, 256, 512},
	xtick=data, x tick label style={font=\footnotesize, yshift=1.5mm}, tick style = transparent,
	ytick={0,0.5,...,2.5},
	y tick label style={font=\footnotesize},
	ymajorgrids=true, major grid style={thin,dashed},
    ylabel={Speedup},
    ylabel style = {font=\scriptsize, yshift=-2mm},
	%axis x line*=bottom,
	%axis y line*=none,
	legend style={draw=none, fill=none, at={(0.48,1.0), font=\scriptsize},
	anchor=south,legend columns=6,
	legend cell align={left},
	/tikz/every odd column/.append style={column sep=1pt},
	/tikz/every even column/.append style={column sep=1pt}},
    legend image code/.code={%
      \draw[#1] (0cm,-0.1cm) rectangle (0.2cm,0.1cm);
    }
	]

\addplot[fill=color2] table [x=id, y=VT] {\datatable};     \addlegendentry{VT}
\addplot[fill=color1] table [x=id, y=Base] {\datatable};\addlegendentry{HyGCN}
\addplot[fill=color7] table [x=id, y=ENGN] {\datatable};\addlegendentry{EnGN}
\addplot[fill=color6] table [x=id, y=AWB] {\datatable};\addlegendentry{AWB-GCN}
\addplot[fill=color8] table [x=id, y=GCNAX] {\datatable};\addlegendentry{GCNAX}
%\addplot[fill=color3] table [x=id, y=FS] {\datatable};  \addlegendentry{\accname{}\textsubscript{Optimal}}
%\addplot[fill=color4] table [x=id, y=FS+VT] {\datatable};  \addlegendentry{\accname{}\textsubscript{Optimal}}
\addplot[fill=color4] table [x=id, y=ATM] {\datatable};  \addlegendentry{\accname{}}
\end{axis}
\end{tikzpicture}  
 \caption{Performance of the \accname for various feature widths.}
 \vspace{-3mm}
 \label{fig:single_f}
  \end{figure}

%% file: eval/scale_new.tex
% \nextgroupplot[%title=\bfseries{100K Items},
% font=\footnotesize\rmfamily,
% title style={yshift=-1.5mm},
% xmajorticks=true,
% ymajorgrids,
% ymax=0.012,
% ymin = 0,
% xtick={4,5,6,7,8,9,10,11,12},
% xlabel = {Learners},
% xlabel shift=-2.5em,
% xlabel near ticks,
% xlabel style={at={(ticklabel cs:0.9)},font =\rmfamily\scriptsize},
% x tick label style = {font =\rmfamily\scriptsize, text width = 1.4cm, align = center, anchor = north},
% ylabel={Time (sec.)},
% ylabel style = {font=\scriptsize},
% ylabel shift=-1.2em,
% y tick label style = {font =\rmfamily\scriptsize, text width = .6cm, align = right, anchor = east,
%  /pgf/number format/fixed,
% },
% legend to name=timelegend,
%   legend style={legend columns=3,font=\rmfamily\scriptsize},
% ]
% \addplot [mark=o,thick,color=blue]table[x=learners,y=FR]{figures/data/learners/100k.txt};
% \addlegendentry{FR}
% \addplot [mark=square,thick,color=olivegreen]table[x=learners,y=Nccl]{figures/data/learners/100k.txt};
% \addlegendentry{Nccl}
% \addplot [mark=x,thick,color=black]table[x=learners,y=BLC]{figures/data/learners/100k.txt};
% \addlegendentry{BLC}

\begin{figure}[t]
    \begin{tikzpicture}
\pgfplotstableread{

%id num dummy	 HBM2   HBM   4xDDR4   DDR4	 HBM2Util   HBMUtil  4xDDR4Util   DDR4Util
% 0	1	1.8	    1.5	    1.3	    1.0	 
% 1	2	2.60	1.95    1.6	    1.2	
% 2   4	3.10	2.3	    1.8	    1.32	
% 3	8	3.5	    2.5	    1.9	    1.34	
% % 4	16	3.7		2.55	1.95	1.35
% 0	1   -2	1.763444774	1.72518015	1.185160625	1 0.4 0.3 0.2 0.1
% 1	2   -2	3.447140831	3.306094595	1.699438106	1.038650939 0.42 0.2 0.22 0.12
% 2	4   -2	6.527702098	5.673892421	2.142722417	1.057958301 0.44 0.34 0.24 0.14
% 3	8   -2	11.18006569	7.162437465	2.555626185	1.065889938 0.46 0.36 0.26 0.16
% 4	16  -2	13.61834339	7.383329051	2.910139341	1.074852157 0.48 0.38 0.28 0.18

% SnF w\LC
id	num	dummy	HBM2	HBM	4xDDR4	DDR4	HBM2Util	HBMUtil	4xDDR4Util	DDR4Util
0	1	-2	1.763444774	1.72518015	1.185160625	1	9.38228125	18.357375	25.25791325	85.28497653
1	2	-2	3.447140831	3.306094595	1.699438106	1.038650939	18.85853125	36.164125	37.18930832	91.04882629
2	4	-2	6.527702098	5.673892421	2.142722417	1.057958301	35.6990625	62.0594375	46.8708558	92.74131455
3	8	-2	11.18006569	7.162437465	2.555626185	1.065889938	61.14375	78.34125	55.92262603	93.4370892
4	16	-2	13.61834339	7.383329051	2.910139341	1.074852157	74.4809375	80.75875	63.73083236	94.22441315
% 5	32	-2	12.67598198	6.923579188	3.118827863	1.075701007	72.1584375 78.296875 70.68792497 94.34460094

% SnF										
% id	num	dummy	HBM2	HBM	4xDDR4	DDR4	HBM2Util	HBMUtil	4xDDR4Util	DDR4Util
% 0	1	-2	3.411972386	3.254544335	2.368328176	1	18.82340625	35.90975	52.37749121	88.47464789
% 1	2	-2	6.458342977	5.441338291	2.743866157	0.9863718871	37.06875	62.470125	63.15169988	90.81173709
% 2	4	-2	10.38350869	6.060918334	2.964519762	0.9694964727	61.25625	71.521875	70.13223916	91.73380282
% 3	8	-2	11.38820972	6.301695451	3.119632359	0.9588104224	68.7178125	76.053125	75.47620164	92.77840376
% 4	16	-2	11.7760716	6.435490428	3.227133705	0.9407429901	72.56125	79.311875	79.72989449	92.94976526
% 5	32	-2	11.94743819	6.51909306	3.301535871	0.9217841429	74.9340625	81.7775	83.02321219	92.70422535

}\datatable
			
\pgfplotscreateplotcyclelist{custom}{%https://tex.stackexchange.com/questions/134346/different-marker-shape-for-pgf-tikz
solid, thick, mark size=3pt, every mark/.append style={solid, thick, fill=color1}, mark=*\\%
dotted, thick, mark size=3pt, every mark/.append style={solid, thick, fill=color2}, mark=square*\\%
densely dotted, thick, mark size=3pt, every mark/.append style={solid, fill=color3}, mark=diamond*\\%
loosely dotted, thick, mark size=3pt, every mark/.append style={solid, fill=color5}, mark=triangle*\\%
solid, thick, mark size=3pt, every mark/.append style={solid, thick, fill=gray}, mark=*\\%
dashed, thick, every mark/.append style={solid, fill=color5},mark=otimes*\\%
loosely dashed, thick, every mark/.append style={solid, fill=gray},mark=*\\%
densely dashed, thick, every mark/.append style={solid, fill=gray},mark=square*\\%
dashdotted, thick, every mark/.append style={solid, fill=gray},mark=otimes*\\%
dashdotdotted, thick, every mark/.append style={solid},mark=star\\%
densely dashdotted,thick, every mark/.append style={solid, fill=gray},mark=diamond*\\%
}

\begin{axis}[
ybar,
bar width=.4em,
%font=\footnotesize\rmfamily,
%title style={yshift=-1.5mm},
axis y line*=left, % dual axis plot https://tex.stackexchange.com/questions/198997/pgfplots-two-y-axis-with-three-plots-and-one-legend
width=.93\columnwidth, 
height=3cm,
enlarge x limits=.13,
xmajorticks=true,
ymajorgrids,
ymax=15,
ymin = 0,
ytick={5,10,15,20},
xtick={0,1,2,3,4,5},
xticklabels={1,2,4,8,16,32},
xlabel = {\#Engines},
xlabel shift=-.5em,
%xlabel near ticks,
xlabel style={at={(ticklabel cs:0.5)},font =\scriptsize},
x tick label style = {font =\scriptsize, text width = 1.4cm, align = center, anchor = north},
ylabel={Speedup},
ylabel style = {font=\scriptsize, yshift=0cm},
y tick label style = {font =\scriptsize, anchor = east, 
 /pgf/number format/fixed, %xshift=.6cm
},
%cycle list name=custom,
legend cell align={left},
legend style={draw=none, fill=none, at={(0.67,1.0), font=\scriptsize},
anchor=south,legend columns=2,
/tikz/every even column/.append style={column sep=0.5cm}},
% legend image code/.code={%
%   \draw[#1] (0cm,-0.1cm) rectangle (0.2cm,0.1cm);
% }
	]  

\addplot[fill=gray] table [x=id, y=dummy] {\datatable};\addlegendentry{\ Speedup}
\addplot[fill=color1] table [x=id, y=DDR4] {\datatable};%\addlegendentry{DDR4-2666 (21.3GB/s)}
\addplot[fill=color2] table [x=id, y=4xDDR4] {\datatable};%\addlegendentry{4 $\times$ DDR4-2666 (85.3 GB/s)}
\addplot[fill=color3] table [x=id, y=HBM] {\datatable};%\addlegendentry{HBM (128GB/s)}
\addplot[fill=color5] table [x=id, y=HBM2] {\datatable};%\addlegendentry{HBM2 (256GB/s)}

\end{axis}

\begin{axis}[
%font=\footnotesize\rmfamily,
%title style={yshift=-1.5mm},
axis y line*=right, % dual axis plot
width=0.94\columnwidth, 
height=3cm,
xmajorticks=false,
%ymajorgrids,
ymax=100,
ymin = 0,
ytick={25,50,75,100},
%xtick={0,1,2,3,4},
%xticklabels={1,2,4,8,16},
%xlabel = {\#Engines},
%xlabel shift=-2.5em,
%xlabel near ticks,
%xlabel style={at={(ticklabel cs:0.9)},font =\rmfamily\scriptsize},
%x tick label style = {font =\scriptsize, text width = 1.4cm, align = center, anchor = north},
ylabel={Mem BW Util. (\%)},
ylabel style = {font=\scriptsize, yshift=0cm},
y tick label style = {font =\scriptsize, anchor = east, 
 /pgf/number format/fixed, xshift=.6cm
},
cycle list name=custom,
legend cell align={left},
legend style={draw=none, fill=none, at={(0.54,1.0), font=\scriptsize},
anchor=south,legend columns=2,
/tikz/every even column/.append style={column sep=0.5cm}},
% legend image code/.code={%
%   \draw[#1] (0cm,-0.1cm) rectangle (0.2cm,0.1cm);
% }
]

\addplot table [x=id, y=DDR4Util] {\datatable};\addlegendentry{DDR4-2666 (21.3GB/s)}
\addplot table [x=id, y=4xDDR4Util] {\datatable};\addlegendentry{4 $\times$ DDR4-2666 (85.3 GB/s)}
\addplot table [x=id, y=HBMUtil] {\datatable};\addlegendentry{HBM (128GB/s)}
\addplot table [x=id, y=HBM2Util] {\datatable};\addlegendentry{HBM2 (256GB/s)}
\addplot table [x=id, y=dummy] {\datatable};\addlegendentry{MemBW Util.}

\end{axis}

\end{tikzpicture}  
 \caption{Sensitivity of \accname to the various memory systems and the number of engines.}
 \label{fig:scale}
  \end{figure}

%% file: eval/cache_size_sensi.tex
   
\begin{figure}[t]
    \begin{tikzpicture}
\pgfplotstableread{

id	data	VT Base	ENGN AWB	GCNAX ATM
0	2	0.6800324532	0.6704597196	0.7010880076	0.8466774159	0.8873305025	1.292134939
1	4	0.8182217707	0.785098849	0.8509795128	0.8351048405	0.9894439865	1.333349329
2	8	0.9163311655	0.8541577166	0.8773528394	1.09763597	1.047006833	1.566315863
3	16	1	0.9047530315	1.060943585	1.268652764	1.110653403	1.743040366
}\datatable

\begin{axis}[
	ybar=0pt,
	enlarge x limits=0.15,
	width=\linewidth, height=3cm,
	bar width=3.5pt, 
	ymin=0,ymax=2.5,
	xticklabels={2MB, 4MB, 8MB, 16MB},
	xtick=data, x tick label style={font=\footnotesize, yshift=1.5mm}, tick style = transparent,
	ytick={0,0.5,...,2.5},
	y tick label style={font=\footnotesize},
	ymajorgrids=true, major grid style={thin,dashed},
    ylabel={Speedup},
    ylabel style = {font=\scriptsize, yshift=-2mm},
	%axis x line*=bottom,
	%axis y line*=none,
	legend style={draw=none, fill=none, at={(0.48,1.0), font=\scriptsize},
	anchor=south,legend columns=6,
	legend cell align={left},
	/tikz/every odd column/.append style={column sep=1pt},
	/tikz/every even column/.append style={column sep=1pt}},
    legend image code/.code={%
      \draw[#1] (0cm,-0.1cm) rectangle (0.2cm,0.1cm);
    }
	]

\addplot[fill=color2] table [x=id, y=VT] {\datatable};     \addlegendentry{VT}
\addplot[fill=color1] table [x=id, y=Base] {\datatable};\addlegendentry{HyGCN}
 \addplot[fill=color7] table [x=id, y=ENGN] {\datatable};\addlegendentry{EnGN}
\addplot[fill=color6] table [x=id, y=AWB] {\datatable};\addlegendentry{AWB-GCN}
\addplot[fill=color8] table [x=id, y=GCNAX] {\datatable};\addlegendentry{GCNAX}
%\addplot[fill=color3] table [x=id, y=FS] {\datatable};  \addlegendentry{\accname{}\textsubscript{Optimal}}
%\addplot[fill=color4] table [x=id, y=FS+VT] {\datatable};  \addlegendentry{\accname{}\textsubscript{Optimal}}
\addplot[fill=color4] table [x=id, y=ATM] {\datatable};  \addlegendentry{\accname{}}
\end{axis}
\end{tikzpicture}  
 \caption{Performance (geomean) of the \accname for various cache sizes.}
 \vspace{-1mm}
 \label{fig:cachesensi}
  \end{figure}

%% file: eval/cachesense.tex
\begin{figure}

\vspace{-2mm}
    \begin{tikzpicture}
\pgfplotstableread{
id	data	Default	Random	RRIP	8ways	32ways	32B	128B	256B
0	PD	1	0.9661687171	1.043751698	0.9977643067	1.001145228	0.7550758516	0.9941666695	0.9633758782
1	CT	1	0.7688748718	1.013134151	0.9990772386	1.000426111	0.6719926242	1.00196986	0.978080734
2	PK	1	0.9507268472	1.037214461	0.9962777088	1.001860532	0.7533183123	0.9528198966	0.8922319761
3	YT	1	0.9131552176	1.002083682	0.9947560243	1.002731915	0.6237266923	0.9995784862	0.8603698289
4	LJ	1	0.8017631033	1.01116794	0.9957668555	1.002171584	0.6138301919	0.982008995	0.924658621
5	OK	1	0.7907040075	0.9997536603	0.9952534186	1.002029945	0.5584699354	0.98471921	0.8980946445
6	RD	1	0.7142041033	1.012134072	0.9940298294	1.002324294	0.4186272136	0.7327653794	0.5146606998
7	Geomean	1	0.8387399621	1.016914487	0.9961308661	1.001812541	0.6174481474	0.944848235	0.8456008745

}\datatable

\pgfplotscreateplotcyclelist{custom}{%https://tex.stackexchange.com/questions/134346/different-marker-shape-for-pgf-tikz
solid, thick, mark size=3pt, every mark/.append style={solid, thick, fill=color1}, mark=*\\%
dotted, thick, mark size=3pt, every mark/.append style={solid, thick, fill=color2}, mark=square*\\%
densely dotted, thick, mark size=3pt, every mark/.append style={solid, fill=color3}, mark=diamond*\\%
loosely dotted, thick, mark size=3pt, every mark/.append style={solid, fill=color5}, mark=triangle*\\%
solid, thick, mark size=3pt, every mark/.append style={solid, thick, fill=gray}, mark=*\\%
dashed, thick, every mark/.append style={solid, fill=color5},mark=otimes*\\%
loosely dashed, thick, every mark/.append style={solid, fill=gray},mark=*\\%
densely dashed, thick, every mark/.append style={solid, fill=gray},mark=square*\\%
dashdotted, thick, every mark/.append style={solid, fill=gray},mark=otimes*\\%
dashdotdotted, thick, every mark/.append style={solid},mark=star\\%
densely dashdotted,thick, every mark/.append style={solid, fill=gray},mark=diamond*\\%
}

\begin{axis}[
	ybar=0pt, %this value determines the space between bars
	width=\linewidth, height=3cm,
	bar width=2.5pt, 
	ymin=0,ymax=1.2,
	xticklabels={PD, CT, PK, YT, LJ, OK, RD, Gmean},
	xtick=data, x tick label style={font=\footnotesize, yshift=1.5mm}, tick style = transparent,
	ytick={0,0.5,...,2.5},
	y tick label style={font=\footnotesize},
	ymajorgrids=true, major grid style={thin,dashed},
    ylabel={Speedup},
    ylabel style = {font=\scriptsize, yshift=-1mm},
	%axis x line*=bottom,
	%axis y line*=none,
	%legend style={draw=none, fill=none, at={(0.69,0.5), font=\tiny},
	legend cell align={left},
	legend style={draw=none, fill=none, at={(0.48,1.0), font=\scriptsize},
	anchor=south,legend columns=4,
	/tikz/every even column/.append style={column sep=0.5cm}},
    legend image code/.code={%
      \draw[#1] (0cm,-0.1cm) rectangle (0.2cm,0.1cm);
    }
	]  

\addplot[fill=color4] table [x=id, y=Default] {\datatable};       \addlegendentry{Default}
\addplot[fill=color6] table [x=id, y=Random] {\datatable};  \addlegendentry{Random}
\addplot[fill=color8] table [x=id, y=RRIP] {\datatable};  \addlegendentry{RRIP}
\addplot[fill=color2] table [x=id, y=8ways] {\datatable};  \addlegendentry{8-ways}
\addplot[fill=color1] table [x=id, y=32ways] {\datatable};\addlegendentry{32-ways}
\addplot[fill=color3] table [x=id, y=32B] {\datatable};  \addlegendentry{32B block}
\addplot[fill=color5] table [x=id, y=128B] {\datatable};  \addlegendentry{128B block}
\addplot[fill=color7] table [x=id, y=256B] {\datatable};  \addlegendentry{256B block}

\end{axis}
\end{tikzpicture}

\vspace{-3mm}

 \caption{\rev{Sensitivity to various cache configurations.}}
 \label{fig:cachesense}
 \vspace{-3mm}
  \end{figure}

%% file: eval/single_sense.tex
\begin{figure}

\centering
\begin{tikzpicture}
\pgfplotstableread{
id	Data	VT	FSVT	VT_Topology	VT_Feature	FSVT_Topology	FSVT_Feature
0	1	80.71	80.71	0.24	24.94	0.24	24.94
1	2	78.50	74.37	0.25	25.23	0.48	23.09
2	4	73.29	65.78	0.28	25.09	0.96	20.58
3	8	64.59	53.23	0.33	24.62	1.92	16.91
4	16	51.31	35.43	0.42	23.52	3.85	11.71
5	32	32.81	22.60	0.62	21.38	4.04	8.91
6	64	10.49	5.20	1.01	18.16	4.43	5.20
7	128	4.96	4.96	1.79	19.49	5.21	7.21

}\datatable

\hspace{-4mm}
%\begin{groupplot}[group style={vertical sep=1em,horizontal sep=3em,group size= 3 by 2},height=3.5cm,width=0.33\textwidth]
\begin{groupplot}[group style={vertical sep=2.8em,horizontal sep=3em,group size= 2 by 1},height=3.6cm,width=0.58\columnwidth]

%(a) 100K items
\nextgroupplot[%title=\bfseries{100K Items},
font=\footnotesize,
title style={yshift=-1.5mm},
height=3cm,
xmajorticks=true,
ymajorgrids,
ymax=100,
ymin = 0,
xtick={0,1,2,3,4,5,6,7},
xticklabels={1,2,4,8,16,32,64,128},
xlabel = {$B_F \times B_V$},
xlabel shift=-.5em,
xlabel near ticks,
xlabel style={at={(ticklabel cs:0.5)},font =\scriptsize},
x tick label style = {font =\scriptsize, text width = 1.4cm, align = center, anchor = north},
ylabel={Cache Miss (\%)},
ylabel style = {font=\scriptsize},
ylabel shift=-1.2em,
y tick label style = {font =\scriptsize, text width = .6cm, align = right, anchor = east,
 /pgf/number format/fixed,
},
	legend style={draw=none, fill=none, at={(0.28,1.1)}, font=\scriptsize,
	anchor=south,legend columns=4,
	/tikz/every column/.append style={column sep=15pt}},
]
\addplot [mark=o,thick,color=blue]table[x=id,y=VT]{\datatable};
\addlegendentry{VT}
\addplot [mark=square,thick,color=olivegreen]table[x=id,y=FSVT]{\datatable};
\addlegendentry{SnF}
%\addplot [mark=x,thick,color=black]table[x=learners,y=BLC]{figures/data/learners/100k.txt};
%\addlegendentry{BLC}

\nextgroupplot[%title=\bfseries{100K Items},
ybar stacked,
bar width=3.5pt, 
font=\footnotesize,
title style={yshift=-1.5mm},
height=3cm,
xmajorticks=true,
ymajorgrids,
ymax=30,
ymin = 0,
xtick={0,1,2,3,4,5,6,7},
xticklabels={1,2,4,8,16,32,64,128},
xlabel = {$B_F \times B_V$},
xlabel shift=-.5em,
xlabel near ticks,
xlabel style={at={(ticklabel cs:0.5)},font =\scriptsize},
x tick label style = {font =\scriptsize, text width = 1.4cm, align = center, anchor = north},
ylabel={Mem. Acc.s (GB)},
ylabel style = {font=\scriptsize},
ylabel shift=-1.2em,
y tick label style = {font =\scriptsize, text width = .6cm, align = right, anchor = east,
 /pgf/number format/fixed,
},
	legend style={draw=none, fill=none, at={(0.28,1.1)}, font=\scriptsize,
	anchor=south,legend columns=4,
	/tikz/every column/.append style={column sep=15pt}},
    legend image code/.code={%
      \draw[#1] (0cm,-0.1cm) rectangle (0.2cm,0.1cm);
    }
]

\addplot[fill=color3,
        postaction={pattern color=black, pattern=north east lines}, 
        xshift=-3.2pt] table [x=id, y=VT_Topology ] \datatable; \addlegendentry{VT Topo.}
\addplot[fill=color1,
        postaction={pattern color=black, pattern=north east lines}, 
        xshift=-3.2pt] table [x=id, y=VT_Feature ] \datatable; \addlegendentry{VT Feat.}

\resetstackedplots{8}
\addplot[fill=color3, xshift=1.6pt] table [x=id, y=FSVT_Topology ] \datatable; \addlegendentry{SnF Topo.}
\addplot[fill=color1, xshift=1.6pt] table [x=id, y=FSVT_Feature] \datatable; \addlegendentry{SnF Feat.}

\end{groupplot}
 % \node[right=1em,inner sep=0pt] at (rel axis cs:0.6,1.25) {\pgfplotslegendfromname{timelegend}};
  
 %   \node[right=1em,inner sep=0pt] at (rel axis cs:1.7,1.25) {\pgfplotslegendfromname{call_legend}};

\end{tikzpicture}
\vspace{-6mm}
  \caption{Miss rate (left) and memory access (right) according to $B_F\times B_V$.}
  \vspace{-3mm}
\label{fig:single_hit_ratio}
\end{figure}

%% file: eval/multi_chip_perf.tex
%\begin{figure}[tp]

%\vspace{-2mm}
\begin{figure}[t]
    \begin{tikzpicture}%[framed,background rectangle/.style={ultra thick,draw=olivegreen}]

%fake engn values for pk and lj and geomean			
\pgfplotstableread{
 id	data	VT HyGCN GCNAX EnGN AWB SAF LAC
0	PD	1	1	1.209277463	1.03053909	1.612479704	1.100597493	2.286663564
1	CT	1	1	1.007044026	1.002011494	1.045404293	1.037773172	1.070468634
2	PK	1	1	1.181124235	1.011083824	1.329251189	1.092419656	1.567977713
3	YT	1	1	0.9734629406	1.014910689	1.112914793	1.02876245	1.134779548
4	LJ	1	1	1.184730434	1.006532439	1.27254502	1.061572456	1.423535093
5	OK	1	0.9980779875	1.194672535	1.058612517	1.291262042	1.124715114	1.843207311
6	RD	1	0.5109004869	1.145346697	1.000217316	1.016427331	1.092419656	1.567977713
7	Gmean	1	0.9076724216	1.083544002	1.050617566	1.148838556	1.070238378	1.453653793

}\datatable

\begin{axis}[
	ybar=0pt, %this value determines the space between bars
	width=\columnwidth, height=3cm,
	bar width=2.5pt, 
	ymin=0,ymax=3,
	xticklabels={PD, CT, PK, YT, LJ, OK, RD, Gmean},
	xtick=data, x tick label style={font=\footnotesize, yshift=1.5mm}, tick style = transparent,
	ytick={0,1,2},
	y tick label style={font=\footnotesize},
	ymajorgrids=true, major grid style={thin,dashed},
    ylabel={Speedup},
    ylabel style = {font=\scriptsize, yshift=-1mm},
	%axis x line*=bottom,
	%axis y line*=none,
	%legend style={draw=none, fill=none, at={(0.69,0.5), font=\tiny},
	legend cell align={left},
	legend style={draw=none, fill=none, at={(0.48,1.0), font=\scriptsize},
	anchor=south,legend columns=6,
	/tikz/every even column/.append style={column sep=0.15cm}},
    legend image code/.code={%
      \draw[#1] (0cm,-0.1cm) rectangle (0.2cm,0.1cm);
    }
	]  

\addplot[fill=color2] table [x=id, y=VT] {\datatable};       \addlegendentry{VT}
\addplot[fill=color1] table [x=id, y=HyGCN] {\datatable};  \addlegendentry{HyGCN}
\addplot[fill=color7] table [x=id, y=EnGN] {\datatable};  \addlegendentry{EnGN}
\addplot[fill=color6] table [x=id, y=AWB] {\datatable};  \addlegendentry{AWB-GCN}
\addplot[fill=color8] table [x=id, y=GCNAX] {\datatable};\addlegendentry{GCNAX}
%\addplot[fill=color3] table [x=id, y=SAF] {\datatable};  \addlegendentry{\accname}
\addplot[fill=color4] table [x=id, y=LAC] {\datatable};  \addlegendentry{\accnameplus} %+LC

\end{axis}
\end{tikzpicture}

\caption{Performance of the multi-chip accelerator.}
\label{fig:multi_chip_perf}
\end{figure}

%% file: eval/reproduce.tex
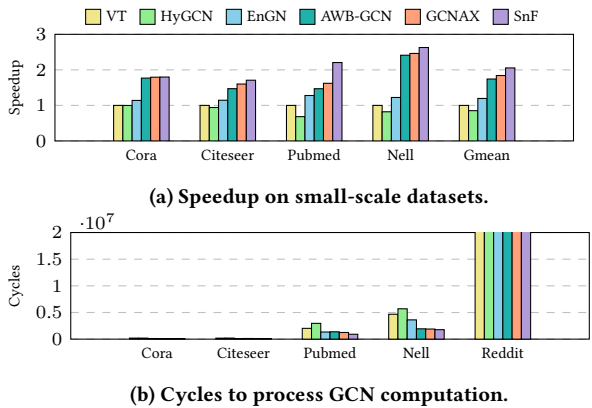
\begin{figure}[t]

\pgfplotsset{
every axis title/.append style={at={(0.5,-0.7)},font=\footnotesize}
}
\pgfplotstableread{
id	data	Base	VT	ATM	AWB GCNAX ENGN
0	Cora	192007	191715	106627	108406	106881	119774
1	Citeseer	197758	186066	108889	126505	116203	111440
2	Pubmed	2963234	2024864	917812	1378040	1248143	1339171
3	Nell	5692890	4670336	1776747	1936122	1896788	3613792
4	Reddit	2476659415	2395348120	600350941	1094464248	847032493	1992095251
}\gnnstableA

% which one is 512?
% table B: 512K...
\pgfplotstableread{
id	data	Base	VT	ATM	AWB GCNAX ENGN
0	Cora	0.9984792221	1	1.797996755	1.768490674	1.793723861	1.140434366
1	Citeseer	0.940877304	1.000000000	1.708770030	1.470818456	1.601213893	1.145317348
2	Pubmed	0.683329043	1.000000000	2.206184827	1.469379858	1.622301523	1.278275702
3	Nell	0.8203805097	1	2.628588088	2.412211627	2.46223405	1.223285062
4	Geomean	0.8518817404	1	2.054514628	1.742520248	1.840417953	1.195465604
}\gnnstableB

\begin{subfigure}[t]{\columnwidth}
    \begin{tikzpicture}
    %[framed,background rectangle/.style={ultra thick,draw=olivegreen}]

\begin{axis}
[	width=\columnwidth,
    ybar=0pt, %this value determines the space between bars
	enlarge x limits=.25,
	height=3cm,
	bar width=3.5pt, 
	ymin=0,ymax=3,
	xticklabels={Cora, Citeseer, Pubmed, Nell, Gmean},
	xtick=data, x tick label style={font=\scriptsize, yshift=1.5mm}, tick style = transparent,
	ytick={0,1,2,3,4,5},
	y tick label style={font=\footnotesize},
	ymajorgrids=true, major grid style={thin,dashed},
    ylabel={Speedup},
    ylabel style = {font=\scriptsize, yshift=-.5mm},
	%axis x line*=bottom,
	%axis y line*=none,
	%legend style={draw=none, fill=none, at={(0.69,0.5), font=\tiny},
	legend style={draw=none, fill=none, at={(0.5,1.0), font=\scriptsize},
	anchor=south,legend columns=8,
	/tikz/every even column/.append style={column sep=.1cm}}, %even: test, odd: square box
    legend image code/.code={%
      \draw[#1] (0cm,-0.1cm) rectangle (0.2cm,0.1cm);
    }
	]

\addplot[fill=color2] table [x=id, y=VT] {\gnnstableB};       \addlegendentry{VT}
\addplot[fill=color1] table [x=id, y=Base] {\gnnstableB};  \addlegendentry{HyGCN}
 \addplot[fill=color7] table [x=id, y=ENGN] {\gnnstableB};  \addlegendentry{EnGN}
\addplot[fill=color6] table [x=id, y=AWB] {\gnnstableB};  \addlegendentry{AWB-GCN}
\addplot[fill=color8] table [x=id, y=GCNAX] {\gnnstableB};\addlegendentry{GCNAX}
% \addplot[fill=color3] table [x=id, y=SAF] {\datatable};  \addlegendentry{\accname}
\addplot[fill=color4] table [x=id, y=ATM] {\gnnstableB};  \addlegendentry{\accnameplus}
\end{axis}
\end{tikzpicture}
\caption{Speedup on small-scale datasets.}
\label{fig:reproduce_speed}
\end{subfigure}

\begin{subfigure}[t]{\columnwidth}
\begin{tikzpicture}
\begin{axis}
[	width=\columnwidth,
    ybar=0pt, %this value determines the space between bars
	enlarge x limits=.25,
	 height=3cm,
	bar width=3.5pt, 
	ymin=0,ymax=20000000,
% 	yscale=log,
	xticklabels={Cora, Citeseer, Pubmed, Nell, Reddit},
	xtick=data, x tick label style={font=\scriptsize, yshift=1.5mm}, tick style = transparent,
	y tick label style={font=\footnotesize},
	ymajorgrids=true, major grid style={thin,dashed},
    ylabel={Cycles},
    ylabel style = {font=\scriptsize, yshift=-.5mm},
	%axis x line*=bottom,
	%axis y line*=none,
	%legend style={draw=none, fill=none, at={(0.69,0.5), font=\tiny},
	legend style={draw=none, fill=none, at={(1.1,1.0), font=\scriptsize},
	anchor=south,legend columns=8,
	/tikz/every even column/.append style={column sep=.5cm}},
    legend image code/.code={%
      \draw[#1] (0cm,-0.1cm) rectangle (0.2cm,0.1cm);
    }
	]
\addplot[fill=color2] table [x=id, y=VT] {\gnnstableA};     %\addlegendentry{VT}
\addplot[fill=color1] table [x=id, y=Base] {\gnnstableA};%\addlegendentry{HyGCN}
 \addplot[fill=color7] table [x=id, y=ENGN] {\gnnstableA};%\addlegendentry{EnGN}
\addplot[fill=color6] table [x=id, y=AWB] {\gnnstableA};%\addlegendentry{AWB-GCN}
\addplot[fill=color8] table [x=id, y=GCNAX] {\gnnstableA};%\addlegendentry{GCNAX}
%\addplot[fill=color3] table [x=id, y=FS] {\datatable};  \addlegendentry{\accname{}\textsubscript{Optimal}}
% \addplot[fill=color4] table [x=id, y=FS+VT] {\datatable};  \addlegendentry{\accname{}\textsubscript{Optimal}}
\addplot[fill=color4] table [x=id, y=ATM] {\gnnstableA};\label{fig:reproduce_cycle}  %\addlegendentry{\accname{}}
\end{axis}
\end{tikzpicture}
\caption{Cycles to process GCN computation.}
\label{fig:reproduce_cycle}
\end{subfigure}

 \caption{Results on small-scale datasets.}
  \vspace{-5mm}
 \label{fig:reproduce}
  \end{figure}
  
  